\documentclass{article}

\PassOptionsToPackage{numbers,sort&compress,square}{natbib}

\usepackage[preprint]{neurips_data_2024}

\usepackage[utf8]{inputenc} %
\usepackage[T1]{fontenc}    %
\usepackage{hyperref}       %
\usepackage{url}            %
\usepackage{booktabs}       %
\usepackage{amsfonts}       %
\usepackage{nicefrac}       %
\usepackage{microtype}      %
\usepackage{xcolor}         %
\usepackage{xspace}
\usepackage{minitoc}
\usepackage{graphicx}
\usepackage{subcaption}
\usepackage{xcolor,soul}
\usepackage{tcolorbox}
\usepackage{refstyle}
\usepackage{cleveref}

\usepackage{colortbl}

\crefformat{section}{\S#2#1#3}
\crefformat{subsection}{\S#2#1#3}
\crefformat{subsubsection}{\S#2#1#3}
\crefrangeformat{section}{\S#3#1#4 to~\S#5#2#6}
\crefmultiformat{section}{\S#2#1#3}{ and~\S#2#1#3}{, #2#1#3}{ and~#2#1#3}
\Crefformat{figure}{#2Figure~#1#3}
\Crefmultiformat{figure}{Figures~#2#1#3}{ and~#2#1#3}{, #2#1#3}{ and~#2#1#3}
\Crefformat{table}{#2Table~#1#3}
\Crefmultiformat{table}{Tables~#2#1#3}{ and~#2#1#3}{, #2#1#3}{ and~#2#1#3}
\Crefformat{appendix}{#2Appendix~\S#1#3}
\Crefmultiformat{appendix}{Appendix~#2#1#3}{ and~#2#1#3}{, #2#1#3}{ and~#2#1#3}
\crefformat{algorithm}{Alg.~#2#1#3}
\Crefformat{equation}{#2Eq.~#1#3}

\newcommand{\stitle}[1]{\vspace{1ex} \noindent{\bf #1.}}
\newcommand{\BENCHMARK}{\mbox{\textsc{MuirBench}}\xspace}
\newcommand{\BENCHMARKexplain}{\textsc{\underline{Mu}lti-\underline{i}mage unde\underline{r}standing \underline{Bench}mark}\xspace}

\newcommand{\grounding}{{visual grounding}\xspace}
\newcommand{\matching}{{image-text matching}\xspace}
\newcommand{\ordering}{{ordering}\xspace}
\newcommand{\scene}{{scene understanding}\xspace}

\newcommand{\cartoon}{{cartoon understanding}\xspace}
\newcommand{\diagram}{{diagram understanding}\xspace}
\newcommand{\geographic}{{geographic understanding}\xspace}
\newcommand{\attributesimilarity}{{attribute similarity}\xspace}
\newcommand{\visualretrieval}{{visual retrieval}\xspace}

\newcommand{\countingshort}{{Counting}\xspace}
\newcommand{\actionshort}{{Action.}\xspace}
\newcommand{\groundingshort}{{Grounding.}\xspace}
\newcommand{\matchingshort}{{Matching.}\xspace}
\newcommand{\orderingshort}{{Ordering}\xspace}
\newcommand{\sceneshort}{{Scene.}\xspace}
\newcommand{\differenceshort}{{Difference.}\xspace}
\newcommand{\cartoonshort}{{Cartoon.}\xspace}
\newcommand{\diagramshort}{{Diagram.}\xspace}
\newcommand{\geographicshort}{{Geographic.}\xspace}
\newcommand{\attributesimilarityshort}{{Attribute.}\xspace}
\newcommand{\visualretrievalshort}{{Retrieval.}\xspace}

\title{\includegraphics[scale=0.05]{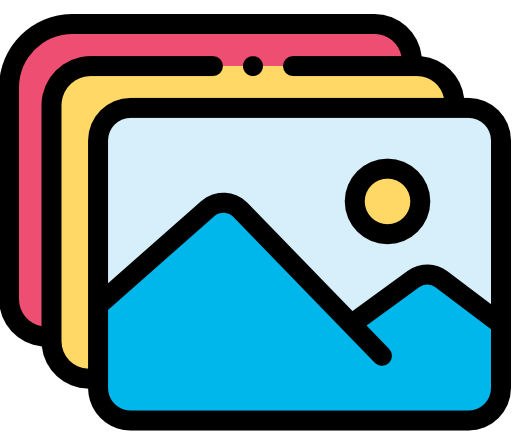} \BENCHMARK: A Comprehensive Benchmark for Robust Multi-image Understanding}

\author{
\textnormal{Fei Wang}$^1$\thanks{Equal leadership. Correspondance to <fwang598@usc.edu; xingyuf2@seas.upenn.edu>.~~~\\ \hspace*{1.1em} $^\dagger$Equal contribution; alphabetic order.~~~}~~
\textnormal{Xingyu Fu}$^2$$^*$~~
\textnormal{James Y. Huang}$^1$$^\dagger$~~
\textnormal{Zekun Li}$^3$$^\dagger$~~
\textnormal{Qin Liu}$^4$$^\dagger$~~
\textnormal{Xiaogeng Liu}$^5$$^\dagger$~~\\
\textnormal{Mingyu Derek Ma}$^6$$^\dagger$~~
\textnormal{Nan Xu}$^1$$^\dagger$~~
\textnormal{Wenxuan Zhou}$^1$$^\dagger$~~
\textnormal{Kai Zhang}$^7$~~
\textnormal{Tianyi Yan}$^1$~~
\textnormal{Wenjie Mo}$^1$~~ \\
\textnormal{Hsiang-Hui Liu}$^3$~~
\textnormal{Pan Lu}$^6$~~ 
\textnormal{Chunyuan Li}$^8$~~ 
\textnormal{Chaowei Xiao}$^5$~~
\textnormal{Kai-Wei Chang}$^6$~~
\textnormal{Dan Roth}$^2$~~\\
\textnormal{Sheng Zhang}$^8$~~
\textnormal{Hoifung Poon}$^8$~~
\textnormal{Muhao Chen}$^4$~~ \\
\small
$^1$USC~~
$^2$UPenn~~ 
$^3$UMN~~ 
$^4$UC Davis
$^5$UW–Madison~~
$^6$UCLA~~ 
$^7$OSU~~ 
$^8$Microsoft Research~~ 
\\ 
\includegraphics[scale=0.02]{figures/icon.png}Project page: \url{https://muirbench.github.io/}
}

\begin{document}

\maketitle

\doparttoc 
\faketableofcontents
\vspace{-15pt}
\begin{center}
    \centering
    \captionsetup{type=figure}
    \includegraphics[width=0.95\textwidth,trim={0cm 0 0cm 0},clip]{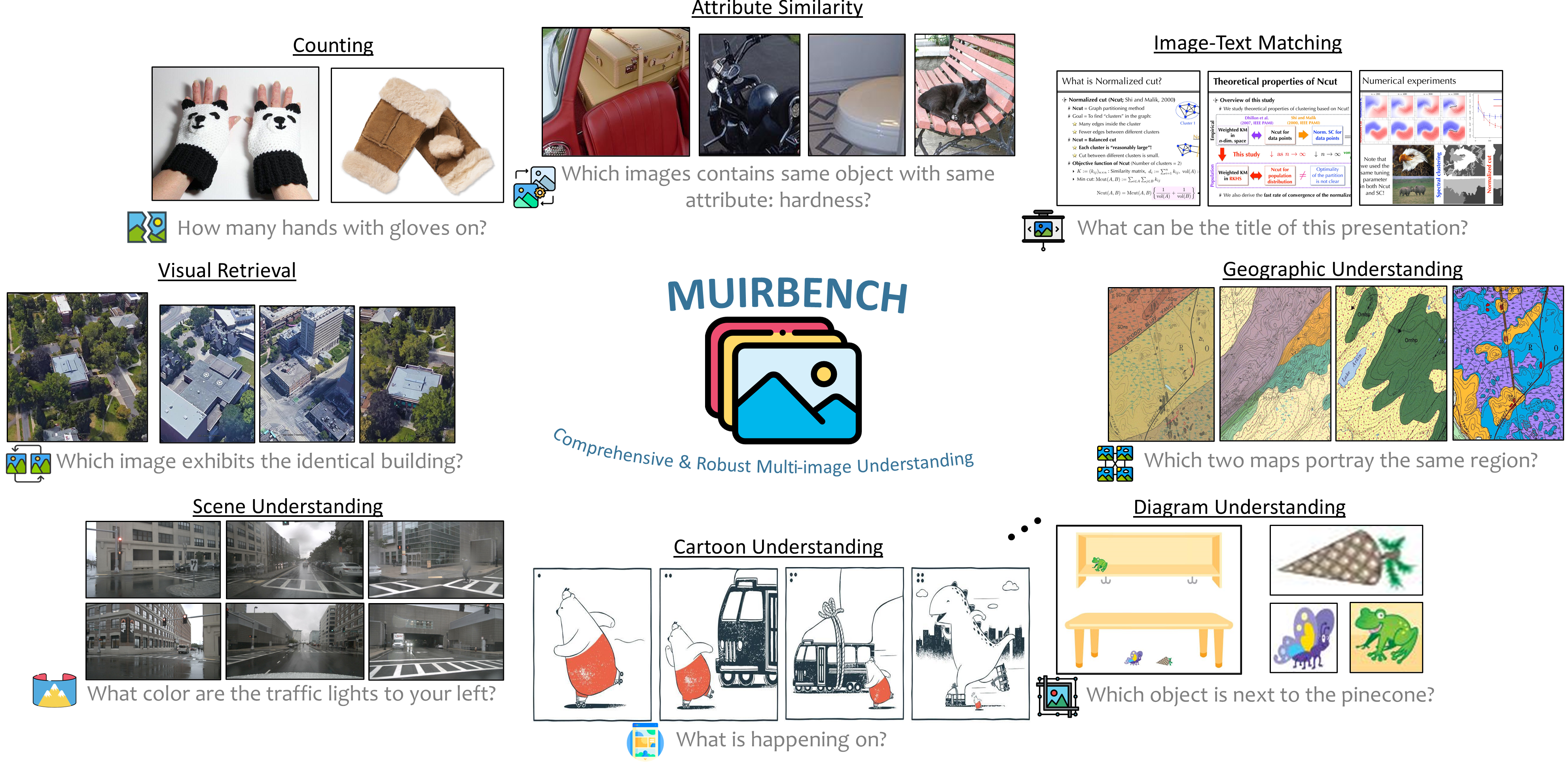}
    \vspace{-0.5em}
\captionof{figure}{{\bf The \BENCHMARK Benchmark.} \BENCHMARK contains 11,264 images and 2,600 multiple-choice questions,
providing robust evaluation on 12 multi-image understanding tasks. Each example comes from one task in \BENCHMARK, presenting diverse multi-image relations.
}

\label{fig:teaser}
\end{center}

\begin{abstract}
We introduce \BENCHMARK, a comprehensive benchmark that focuses on robust multi-image understanding capabilities %
of multimodal LLMs.
\BENCHMARK consists of 12 diverse multi-image tasks (\eg, \scene, \ordering) that involve 10 categories of multi-image relations (\eg, multiview, temporal relations).
Comprising 11,264 images and 2,600 multiple-choice questions,
\BENCHMARK is created in a pairwise manner, where each standard instance is paired with an unanswerable variant that has minimal semantic differences, in order for a reliable assessment. 
Evaluated upon 20 recent multi-modal LLMs, our results reveal that even the best-performing models like GPT-4o and Gemini Pro find it challenging to solve \BENCHMARK, achieving 68.0\% and 49.3\% in accuracy. 
Open-source multimodal LLMs trained on single images can hardly generalize to multi-image questions, hovering below 33.3\% in accuracy.
These results highlight the importance of \BENCHMARK in encouraging the community to develop multimodal LLMs that can look beyond a single image, suggesting potential pathways for future
improvements.

\end{abstract}

\vspace{-3mm}
\section{Introduction}

The proverb ``a picture is worth a thousand words'' is often cited to emphasize the richness of visual information hidden in one image~\cite{gropper1963picture,hibbing2003picture}. However, an image is only a single projection of the real world captured from a specific angle at a specific moment in time~\cite{hays2008im2gps}. In contrast, humans naturally observe multiple images -- multiple pieces of such projections from discrete moments under various scenes -- to perceive and understand the world as a holistic part.
Humans excel at synthesizing information from multiple image sources, whether it involves telling stories from a series of cartoon images~\cite{cohn2017picture,li2023seed}, drawing comparisons among multiple charts and diagrams to infer holistic new insights~\cite{masry-etal-2022-chartqa}, learning from diverse visual experiences such as online lesson slides to adopt new skills~\cite{nouri2005effect}, predicting future event actions from past screenshots~\cite{oh2015action,finn2016unsupervised}, or conducting temporal reasoning based on nuanced differences between photographs~\cite{fu-etal-2022-theres}.
Moreover, multi-image input has the advantage of conveying visuospatial ideas directly -- combining multiple images of the same scene can reveal spatial relations or other more abstract relations in the world ~\cite{faugeras2001geometry}. Multi-image input also overcomes the limitations of resolution that single images face, allowing for better visual perception and understanding~\cite{kawulok2019deep}. 

\begin{figure}
    \centering
    \includegraphics[width=\textwidth]{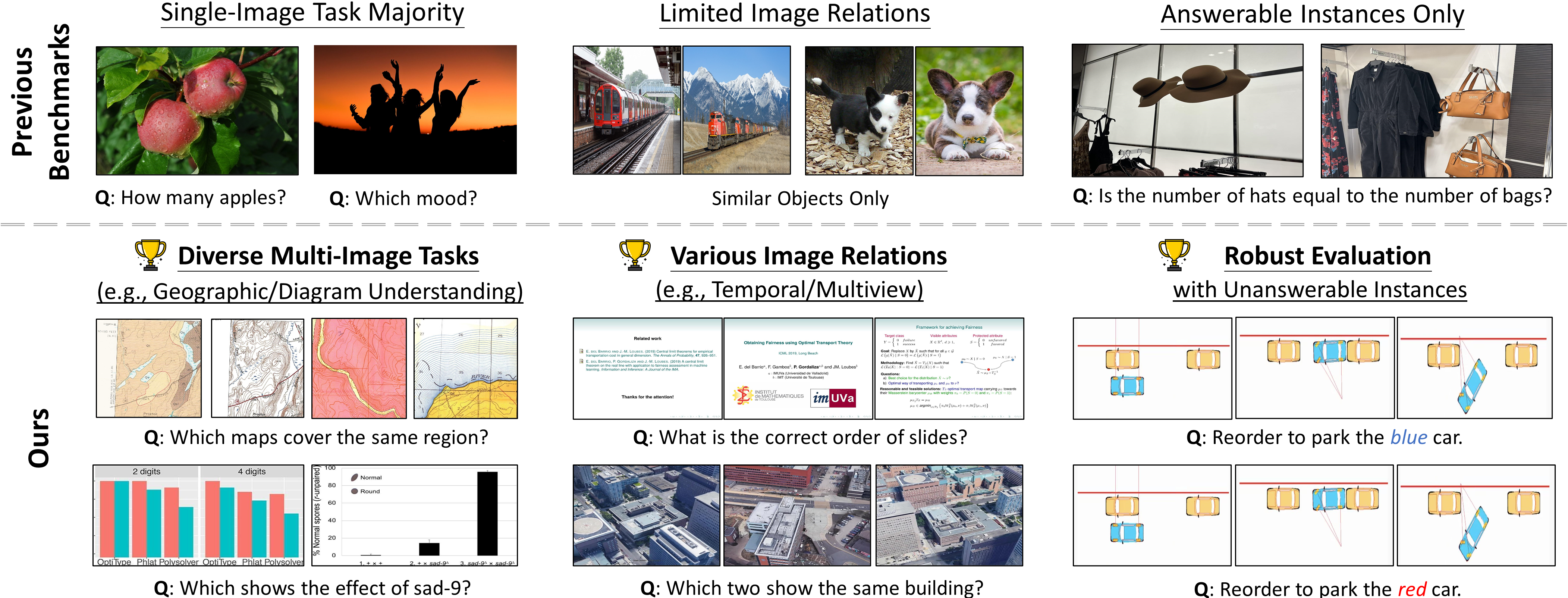}
    \caption{
    Compared with previous benchmarks, \BENCHMARK has several novel features: (1) It evaluates on a comprehensive range of 12 multi-image understanding abilities, \eg ~\geographic and \diagram as introduced in \Cref{sec:bench}, while prior benchmarks generally contain single-image questions. (2) It contains 10 diverse multi-image relations, \eg narrative and complementary as discussed in \Cref{sec:bench}. (3) It provides a robust evaluation on models by unanswerable instance variants. The samples of previous benchmarks are from ~\cite{liu2023mmbench,suhr2019corpus,jiang2024mantis}.
    }
    \label{fig:comparison}
\end{figure}

As multimodal large language models (LLMs)~\cite{gpt4, team2023gemini, liu2023improvedllava, liu2024llavanext, alayrac2022flamingo, bai2023qwenvl, wang2023cogvlm, dai2023instructblip, lu2023unified, zhang2024llamaadapter,Team2024ChameleonME, chen2023internvl, chaves2024training} have begun to show superior performance across various single-image tasks, we now expect them to solve %
hard tasks that require an holistic understanding of multiple images.
This work aims at highlighting crucial aspects of multi-image understanding that have been overlooked when evaluating multimodal LLMs, and providing a comprehensive benchmark for robust multi-image reasoning.
As shown in \autoref{fig:comparison}, current evaluations~\cite{balanced_vqa_v2,liu2023mmbench, li2023seed, yue2023mmmu, lu2023mathvista, liu2023visual, liu2023hidden} generally focus on single-image understanding, thereby neglecting the richer, more complex tasks of integrating and reasoning across multiple images. While many of these benchmarks have been popularized as the de~facto evaluation measures for influential models like GPT-4-Turbo~\cite{gpt4} and Gemini-Pro~\cite{team2023gemini}, this oversight limits the potential of these models to conduct advanced-level multimodal comprehension. 
Though some recent benchmarks start to include multi-image questions in evaluation (\eg, Mantis-Eval~\cite{jiang2024mantis} and BLINK~\cite{fu2024blink}), they are far from being comprehensive in multi-image evaluation that involve multi-persectives, multi-relations and robustness concerns.

In this paper, \textbf{we introduce \BENCHMARK (\BENCHMARKexplain)}, a comprehensive benchmark designed to rigorously assess and evaluate multi-image understanding by multimodal LLMs. \BENCHMARK encompasses 11,264 images and 2,600 multiple-choice questions spanning across 12 distinctive multi-image understanding tasks, \eg \visualretrieval, \cartoon, and \attributesimilarity, \etc.
As illustrated in \autoref{fig:teaser}, there can be multiple images interleaved in the contexts or questions, or presented as choices in our benchmark. 
Instances in \BENCHMARK also contain diverse kinds of multi-image relations, \eg temporal, ordered-pages, or narrative relations, \etc as shown in \autoref{fig:relation_distribution}.
The questions and choices are either derived from the datasets, or manually written by experts.
Additionally, \BENCHMARK adopts a pairwise design approach, where each question-answering instance is paired with a %
expert-annotated
unanswerable counterpart \cite{rajpurkar2018know} featuring minimal differences following \autoref{fig:unanswerable}. 
This design ensures a reliable assessment of multimodal LLMs, mitigating the risk of achieving correct answers through vision or language shortcuts. 
We also include various fine-grained expert annotated labels such as image positions and image types in \BENCHMARK, to facilitate detailed model analysis.

We conduct a comprehensive evaluation on \BENCHMARK using 20 multimodal LLMs of various sizes, including models that %
accept multi-image inputs and those originally designed for single-image inputs.
Experimental results underscore the current limitations of even the most influential multimodal LLMs, \eg GPT-4o and Gemini Pro, in handling multi-image scenarios. For instance, GPT-4o and Gemini Pro achieve mere 68.0\% and 49.3\% of accuracy respectively, which are 25.1 \% and 43.8\% lower than human performance.
We also show that multimodal LLMs perform much worse on unanswerable questions than their answerable counterparts, with GPT-4o and Gemini Pro exhibiting accuracy gaps of 26.8\% and 21.5\%. %
Furthermore, multimodal LLMs trained solely on single images demonstrate impaired generalization to multi-image contexts. %
These findings highlight the significance of \BENCHMARK in driving the development of multimodal LLMs in transcending single-image limitations. 
We believe \BENCHMARK can serve as an effective testbed for holistic multi-image understanding, encouraging the community to cultivate models with a more comprehensive and integrated understanding of the visual world.

\vspace{-3mm}
\section{Related Work}

\subsection{Multimodal Understanding Benchmarks}

A number of recent benchmarks have been developed to comprehensively assess the multimodal understanding and reasoning capabilities of multimodal language models (LLMs)~\cite{lu2021inter,lu2022learn,li2023seed,liu2023mmbench,lu2023mathvista,yue2023mmmu,mathverse2024zhang,ying2024mmt}.
However, most of these benchmarks primarily focus on single-image scenarios.
While some benchmarks include multi-image examples  \cite{lu2023mathvista,yue2023mmmu,fu2024blink,ying2024mmt,wang2024mementos}, they typically require limited aspects of capacities (\eg, image comparison for MathVista) and do not provide a comprehensive assessment of multimodal LLMs in multi-image scenarios.
While some benchmarks feature video understanding \cite{grauman2022ego4d,maaz2023video} or in-context learning \cite{shukor2023beyond,jiang2024many}, the assessed capabilities are fundamentally different from multi-image understanding. 
Video understanding focuses on continuous streams of frames capturing dynamic changes over time, while in-context learning focuses on task adaptation using few-shot examples. 
In contrast, multi-image understanding challenges models to integrate and analyze spatial and contextual cues from varied perspectives, settings, and moments, thereby simulating the way humans process information from multiple visual sources.
Recently, there have been dedicated efforts to assess multimodal LLMs in multi-image scenarios.
For example, MANTIS-Eval~\cite{jiang2024mantis} is a human-annotated benchmark comprising 207 examples for multi-image reasoning, such as size perceptions and weight comparisons.
DEMON~\cite{li2023fine} evaluates whether multimodal LLMs can follow zero-shot demonstrative instructions. 
However, these benchmarks still focus on limited multi-image relations or reasoning processes and lack of robust evaluation.
In contrast, \BENCHMARK provides a comprehensive assessment of multimodal LLMs, covering a broader range of multi-image capacities.

\subsection{Multimodal Large Language Models}

Inspired by the remarkable achievements in recent LLMs~\cite{gpt3,gpt4, touvron2023llama, zheng2024judging,MosaicML2023Introducing}, a series of studies have begun exploring multimodal LLMs that can concurrently interpret visual and linguistic information. However, most of early multimodal LLMs are trained on single-image datasets and overlook the complicated tasks of multi-image understanding \cite{liu2023improvedllava,dai2023instructblip,chen2023minigptv2}. 
Recent work starts training multimodal LLMs on interleaved image-text corpus such as MMC4~\citep{zhu2024multimodal} and OBELICS~\citep{laurenccon2024obelics} for pretraining as well as Mantis-Instruct~\citep{jiang2024mantis} for instruction tuning, which enables models to generate texts given multiple images. 
While some of these models, like Flamingo~\cite{alayrac2022flamingo}, Idefics~\citep{laurenccon2024obelics}, Emu~\cite{sun2023generative}, and VILA~\citep{lin2023vila}, have demonstrated in-context learning capabilities, there is still a lack of evidence regarding their capabilities in understanding multiple images within independent instances. 
Although instruction tuned models such as Mantis~\citep{jiang2024mantis} and GPT-4-Turbo~\citep{gpt4} have shown to possess counting and comparison skills over multi-image inputs, their ability in understanding and reasoning over multiple images with different relations across diverse tasks, though critical, remain unexplored. Therefore, we propose \BENCHMARK to conduct comprehensive evaluation and provide insights to further improve their capabilities in handling realistic multi-image tasks.

\begin{figure}[t]
     \centering
     \begin{minipage}[b]{0.47\textwidth}
        \centering
        \includegraphics[width=\textwidth]{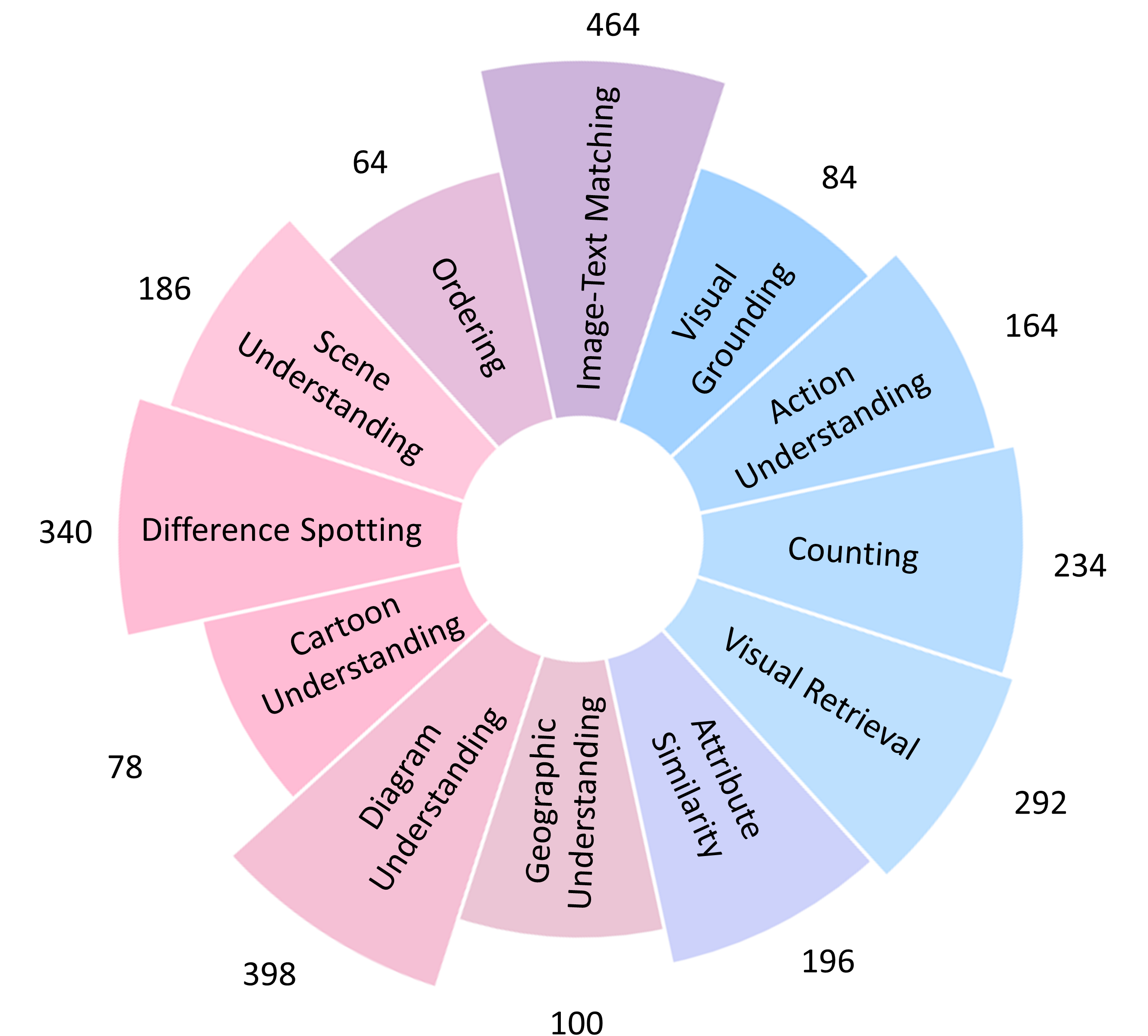}
        \caption{Data distribution by tasks in \BENCHMARK. More details are in \S\ref{sec:bench}.}
        \label{fig:distribution}
     \end{minipage}
     \hfill
     \begin{minipage}[b]{0.5\textwidth}
        \centering
        \includegraphics[width=\textwidth]{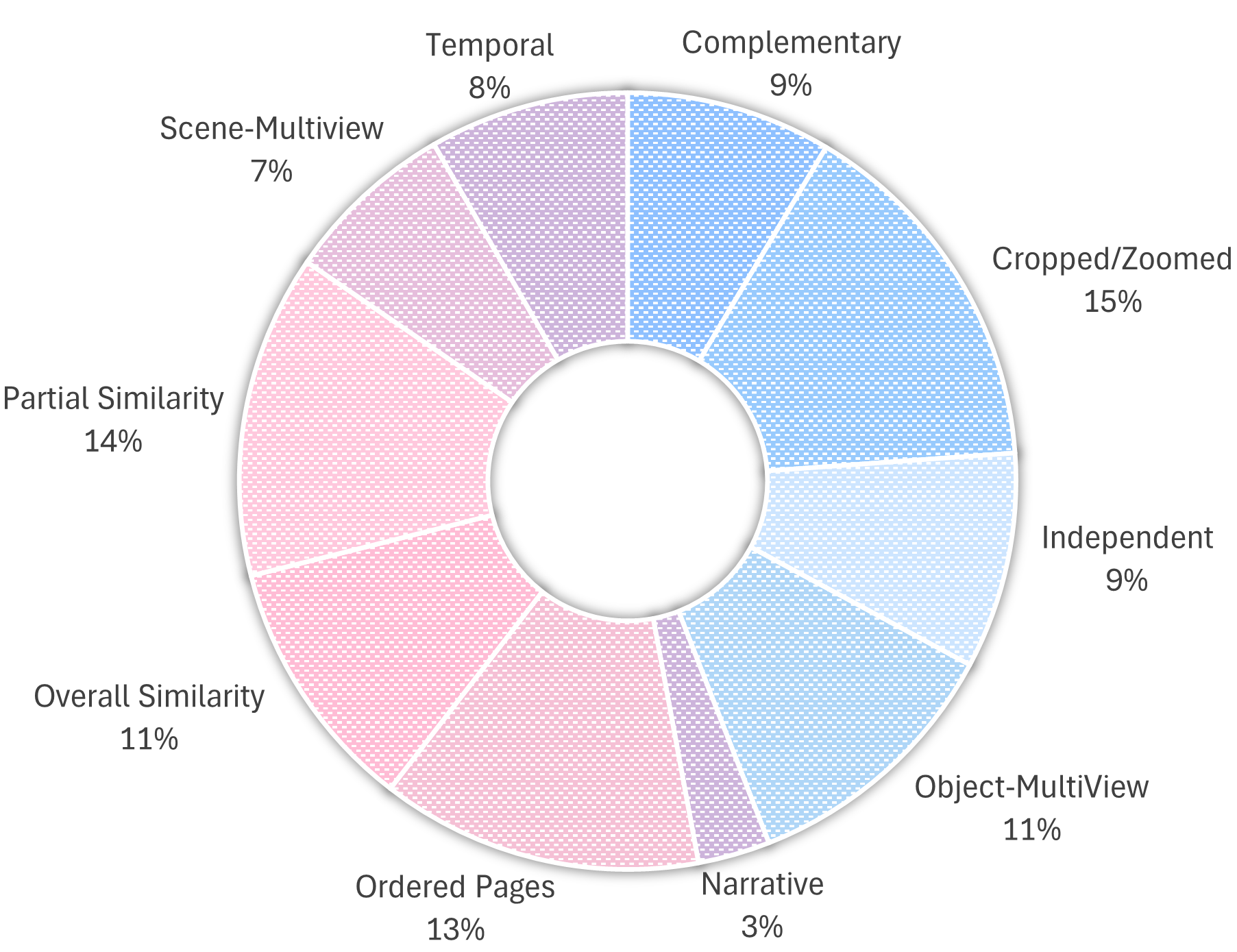}
        \caption{Data distribution by multi-image relation categories. More details are in \S\ref{sec:bench}.}
        \label{fig:relation_distribution}
     \end{minipage}
\end{figure}
\vspace{-3mm}

\section{\BENCHMARK}
\label{sec:bench}

Our benchmark is meticulously curated for comprehensively assessing multimodal LLMs' capabilities %
in holistic multi-image understanding.
We introduce the overall design and key features of \BENCHMARK in \Cref{sec:data_over}, and delve deep into the data curation process in \Cref{sec:data_collect}.

\subsection{Benchmark Overview}
\label{sec:data_over}
Focusing on multi-image understanding, \BENCHMARK consists of 11,264 images and 2,600 multiple-choice questions, with an average of 4.3 images per instance. In general, \BENCHMARK adheres to two key design principles. First, it seeks to provide a \textbf{comprehensive} and holistic evaluation on multimodal LLMs' multi-image understanding capabilities, by containing 12 diverse multi-image tasks covering 10 distinctive multi-image relation categories. Additional fine-grained labels such as input image positions and image types are also included to support comprehensive analysis of models. 
Second, it seeks to provide a \textbf{robust} evaluation, following a pairwise design where each answerable instance is paired with an unanswerable counterpart featuring minimal differences.

\stitle{Comprehensive Multi-Image Evaluation}
\BENCHMARK provides an comprehensive assessment through 12 distinctive multi-image understanding tasks, with selected examples of each task shown in \autoref{fig:results-qual}. 
As illustrated in \autoref{fig:distribution}, each task represents 2.5\% to 17.8\% of the whole benchmark. 

\textsc{[Action Understanding]} aims to evaluate the ability of models to understand continuous images in chronological order and match it with an action.
\textsc{[Attribute Similarity]} aims to evaluate the ability of models to identify a specific given attribute among multiple images.
\textsc{[Cartoon Understanding]} aims to evaluate the ability of models to understand stories conveyed in cartoon images.
\textsc{[Counting]} aims to evaluate the ability of models to count the number of specific objects across multiple images.
\textsc{[Diagram Understanding]} aims to evaluate the ability of models to understand information conveyed in diagram images.
\textsc{[Difference Spotting]} aims to evaluate the ability of models to identify differences across multiple images.
\textsc{[Geographic Understanding]} aims to evaluate the ability of models to understand maps and reason upon geographic features.
\textsc{[Image-text Matching]} aims to evaluate the ability of models to understand the meaning of a text snippet and match it with the corresponding visual content or vice versa.
\textsc{[Ordering]} aims to evaluate the ability of models to order a series of images based on the textual description.
\textsc{[Scene Understanding]} aims to evaluate the ability of models to understand a scene comprised of multiple views from multiple surveillance images.
\textsc{[Visual Grounding]} aims to evaluate the ability of models to ground a specific object and seek information about it within multiple images.
\textsc{[Visual Retrieval]} aims to evaluate the ability of models to retrieval images that contain the same building.

Additionally, \BENCHMARK includes images covering 10 various categories of multi-image relations, such as narrative images conveying stories or ideas, ordered pages of documents and slides providing collective insights, images forming a temporal sequence presenting events, and multiple views of objects or 3D scenes offering a complete vision, with the complete distribution shown in \Cref{fig:relation_distribution}. 
In terms of image presentation, the number of images in each instance ranges from two to nine, while the input positions of  images can be the beginning of question, middle of question, end of question, options, and a mix of these positions.
\BENCHMARK also exhibits various image types, including but not limited to slides, maps, medical images, drone/satellite images, animations, memes, graphics, and 3D views.
The data diversity from the aforementioned perspectives enhances the comprehensiveness of our benchmark.
More details can be found in \autoref{sec:appendix:dataset}.

\begin{figure}[t]
    \centering
    \includegraphics[width=\textwidth]{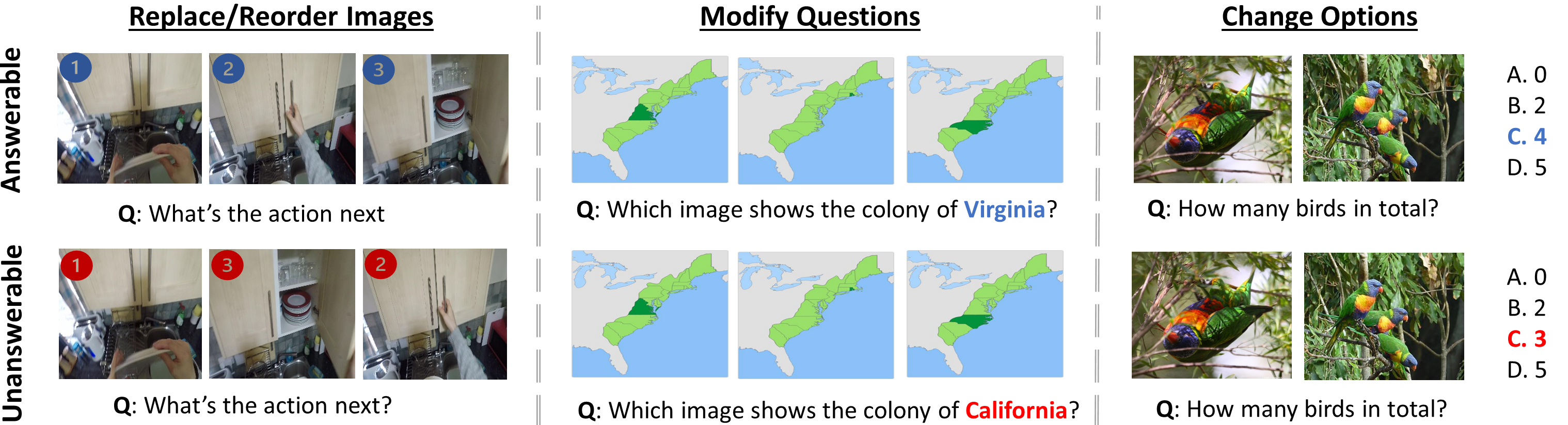}
    \caption{Three major strategies are used in \BENCHMARK to create unanswerable instances from their answerable counterparts with minimal changes (\S\ref{sec:data_collect}). In the above examples, \textcolor{blue}{blue} marks denote the original input in the answerable case, and \textcolor{red}{red} marks highlight the input in the unanswerable case.}
    \label{fig:unanswerable}
\end{figure}

\stitle{Robust Evaluation}
Existing datasets primarily assess models' capabilities in solving answerable questions but overlook their ability to recognize what they do not know \cite{rajpurkar2018know,miyai2024unsolvable}. In real-world scenarios, there is no guarantee that user queries are answerable. A reliable multimodal LLM should directly indicate when a query is unanswerable rather than providing an answer that is most likely to be correct. In light of this, we pair each answerable instance with an unanswerable counterpart, featuring minimal differences, to provide a more robust evaluation, simulating real-world scenarios. 
We adopt multiple strategies to manually design the unanswerable instances, with major strategies of image replacing or reordering, question modification, and option modification introduced in \autoref{fig:unanswerable}.
More details can be found in \autoref{sec:appendix:dataset}.

\subsection{Data Collection}
\label{sec:data_collect}

\stitle{Answerable Data Collection}
We invest our efforts in collecting multi-image multiple-choice question answering (MCQA) data covering various tasks and multi-image relations.
Diverse data attributes enable fine-grained and diagnostic evaluation, while the multiple-choice format ensures deterministic results.
To achieve this goal, we consider three sources of data, including existing datasets, dataset derivations, as well as newly collected data. 
\textit{Existing data} (40.8\%) come from GeneCIS \cite{vaze2023genecis}, SeedBench \cite{li2023seed}, and IconQA \cite{lu2021iconqa}.
\textit{Derived data} (21.7\%) reformat data into MCQA format, using multiple strategies including question generation, option rewriting, and single-image QA combination, \etc upon instances from NLVR2 \cite{suhr2019corpus}, HallusionBench \cite{guan2023hallusionbench}, ISVQA \cite{bansal2020visual}, and MMBench \cite{liu2023mmbench}.
\textit{New data} (37.5\%) address certain tasks (\eg~\geographic and \visualretrieval) that are underrepresented in the aforementioned collection to fulfill a more comprehensive evaluation. We manually create the question and choices for these data based on images from the National Geologic Map Database\footnote{\url{https://ngmdb.usgs.gov/ngmdb/ngmdb\_home.html}}, University-1652 \cite{zheng2020university, zheng2023uavm}, PubMed papers\footnote{\url{https://pubmed.ncbi.nlm.nih.gov/}}, and SciDuet slides~\cite{sun2021d2s}. 
Details about curation process and data sources for each task can be found in \autoref{sec:appendix:dataset}.

\stitle{Unanswerable Data Collection}
As shown in \Cref{fig:unanswerable}, we consider three strategies for modifying an answerable instance to its unanswerable counterpart with minimal changes. 
We first replace or reorder some images to disrupt the question-image and image-image relations (24.2\%). 
We also modify the question to make it incompatible with the images and options (35.3\%). 
In addition, we replace options to create a scenario with no correct answer (40.5\%). 
For each answerable instance, we apply one of these three strategies. 
More details can be found in \autoref{sec:appendix:dataset}.

\stitle{Quality Control}
We employ two types of quality control throughout the annotation process: automatic check with predefined rules, and a manual examination of each instance to filter out any low-quality data. 
The automatic check verifies valid instance format, answers, metadata values, and the coreference between image placeholders and images (ensuring no redundant image), as well as the accessibility of images.
The manual examination is conducted by four experts working in this field, and filters out ambiguous queries, unclear images, and confusing instances.

\section{Experiments}
\label{sec:exp}

In this section, we first describe the experimental setup and the baselines (\Cref{sec:exp_setup}).
Then we present a comprehensive evaluation of 20 recent multimodal LLMs (\Cref{sec:exp_results}). We demonstrate that while humans can answer the questions with high accuracy, \BENCHMARK is challenging for existing models.
Finally, we conduct various analyses on multiple experiment settings, including sensitivity to various resolution and error analysis (\Cref{sec:exp_analysis}).

\begin{table}[t!]
    \centering
    \scalebox{0.75}{{\fontsize{9pt}{12pt}\selectfont
    \begin{tabular}{lccccccc} 
    \toprule[1.2pt] 
    & \begin{tabular}{c} Overall \\$(2,600)$\end{tabular} 
    & \begin{tabular}{c} \countingshort \\$(234)$\end{tabular} 
    & \begin{tabular}{c} \actionshort \\$(164)$\end{tabular}
    & \begin{tabular}{c} \groundingshort \\$(84)$\end{tabular} 
    & \begin{tabular}{c} \matchingshort \\$(464)$\end{tabular} 
    & \begin{tabular}{c} \orderingshort \\$(64)$\end{tabular}
    & \begin{tabular}{c} \sceneshort \\$(186)$\end{tabular}    \\
    \midrule[1.2pt]
    Random Choice & 23.99 & 20.98 & 23.41 & 25.00 & 24.12 & 22.81 & 25.00\\
     Human & 93.15 & 94.87 & 97.56 & 85.71 & 94.83 & 87.50 & 94.62 \\
     \hline 
     \rowcolor{lightgray} 
    \multicolumn{8}{c}{ \textit{\textbf{Multi-Image input multimodal LLMs}}} \\
    \hline 
    GPT-4o~\cite{gpt4} & 68.00 & 49.15 & 44.51 & 36.90 & 86.85 & 23.44 & 71.51\\ 
    GPT-4-Turbo~\cite{gpt4} & 62.31 & 42.31 & 39.63 & 53.57 & 80.39 & 35.94 & 59.14\\ 
    Gemini Pro~\cite{team2023gemini} & 49.35 & 28.63 & 35.98 & 28.57 & 66.59 & 12.50 & 59.14\\ 
    Mantis-8B-Idefics2~\cite{jiang2024mantis}& 44.50 & 38.46 & 33.54 & 26.19 & 53.88 & 18.75 & 56.99\\ 
    Mantis-8B-clip-llama3~\cite{jiang2024mantis} & 37.38 & 29.06 & 36.59 & 21.43 & 43.32 & 18.75 & 56.99\\ 
    Mantis-8B-siglip-llama3~\cite{jiang2024mantis} & 36.12 & 27.35 & 37.20 & 22.62 & 43.75 & 7.81 & 54.30\\ 
    Idefics-9B-Instruct~\cite{laurenccon2024obelics} & 35.43 & 29.91 & 28.05 & 13.10 & 35.99 & 12.50 & 27.41 \\
    Emu2-Chat (37B)~\cite{sun2023generative} & 33.62 & 31.20 & 27.44 & 26.19 & 37.28 & 15.63 & 48.39 \\
    VILA1.5-13B~\cite{lin2023vila} & 33.12 & 19.66 & 28.66 & 25.00 & 40.95 & 10.94 & 56.45\\ 
    Idefics2-8B~\cite{idefics2} & 26.08 & 21.79 & 26.22 & 26.19 & 24.78 & 15.62 & 56.45\\ 
    OpenFlamingo-v2-9B~\cite{awadalla2023openflamingo} & 23.73 & 21.79 & 26.83 & 30.95 & 24.14 & 21.88 & 22.58\\ 
     \hline 
    \rowcolor{lightgray} 
    \multicolumn{8}{c}{ \textit{\textbf{Single-Image input multimodal LLMs}}} \\
    \hline 
    LLaVA-NeXT-34B~\cite{liu2024llavanext} & 33.31 & 36.32 & 26.22 & 33.33 & 37.93 & 21.88 & 54.30\\ 
    LLaVA-v1.5-7B-xtuner~\cite{2023xtuner} & 33.23 & 26.92 & 25.61 & 23.81 & 22.84 & 4.69 & 39.78\\ 
    Yi-VL-6B~\footref{Yifootnote} & 28.69 & 28.21 & 27.44 & 28.57 & 25.00 & 7.81 & 38.71\\ 
    LLaVA-internLM2-7B~\cite{2023internlm} & 28.15 & 34.19 & 26.22 & 32.14 & 25.65 & 7.81 & 42.47\\ 
    LLaVA-v1.5-13B~\cite{liu2023improvedllava} & 24.38 & 25.21 & 29.27 & 14.29 & 20.26 & 20.31 & 36.56\\ 
    LLaVA-v1.5-7B~\cite{liu2023improvedllava} & 23.46 & 23.08 & 27.44 & 14.29 & 23.49 & 23.44 & 34.95\\ 
    LLaVA-v1.5-13B-xtuner~\cite{2023xtuner} & 21.69 & 23.08 & 23.17 & 16.67 & 21.98 & 14.06 & 47.85\\ 
    CogVLM~\cite{wang2023cogvlm} & 20.85 & 14.10 & 26.22 & 16.67 & 21.34 & 12.50 & 41.40\\ 
    MiniGPT-4-v2~\cite{chen2023minigptv2} & 17.35 & 11.97 & 14.02 & 25.00 & 17.03 & 18.75 & 14.52\\ 
    \bottomrule
    \end{tabular}}}
    \vspace{2em}
    \scalebox{0.75}{{\fontsize{9pt}{12pt}\selectfont
    \setlength{\tabcolsep}{7pt}
    \begin{tabular}{lcccccc} 
    \toprule[1.2pt] 
    & \begin{tabular}{c} \differenceshort \\$(340)$\end{tabular} 
    & \begin{tabular}{c} \cartoonshort \\$(78)$\end{tabular}
    & \begin{tabular}{c} \diagramshort \\$(398)$\end{tabular} 
    & \begin{tabular}{c} \geographicshort \\$(100)$\end{tabular} 
    & \begin{tabular}{c} \attributesimilarityshort \\$(196)$\end{tabular}
    & \begin{tabular}{c} \visualretrievalshort \\$(292)$\end{tabular}    \\
    \midrule[1.2pt]
    Random Choice & 23.18 & 25.00 & 29.56 & 25.00 & 20.00 & 21.30 \\
     Human & 92.94 & 82.05 & 98.99 & 98.00 & 87.76 & 86.30 \\
     \hline 
    \rowcolor{lightgray} 
    \multicolumn{7}{c}{ \textit{\textbf{Multi-Image input multimodal LLMs}}} \\
    \hline 
    GPT-4o~\cite{gpt4} & 60.29 & 51.28 & 88.69 & 56.00 & 56.12 & 80.14\\ 
    GPT-4-Turbo~\cite{gpt4} & 60.59 & 52.56 & 79.15 & 57.00 & 50.51 & 64.04\\ 
    Gemini Pro~\cite{team2023gemini} & 45.29 & 47.44 & 64.82 & 48.00 & 41.33 & 43.84\\ 
    Mantis-8B-Idefics2~\cite{jiang2024mantis} & 28.82 & 38.46 & 67.59 & 26.00 & 48.47 & 35.62\\ 
    Mantis-8B-clip-llama3~\cite{jiang2024mantis} & 24.12 & 43.59 & 54.27 & 16.00 & 33.67 & 31.85\\ 
    Mantis-8B-siglip-llama3~\cite{jiang2024mantis} & 27.35 & 46.15 & 47.99 & 22.00 & 31.63 & 28.08\\ 
    Idefics-9B-Instruct~\cite{laurenccon2024obelics} & 34.41 & 48.72 & 46.98 & 35.00 & 32.65 & 43.49 \\
    Emu2-Chat (37B)~\cite{sun2023generative} & 32.65 & 43.59 & 37.69 & 34.00 & 31.63 & 23.97 \\
    VILA1.5-13B~\cite{lin2023vila} & 24.71 & 30.77 & 42.71 & 31.00 & 24.49 & 30.14\\ 
    Idefics2-8B~\cite{idefics2} & 27.65 & 39.74 & 25.38 & 21.00 & 17.86 & 17.12\\ 
    OpenFlamingo-v2-9B~\cite{awadalla2023openflamingo} & 21.76 & 25.64 & 31.91 & 25.00 & 18.88 & 15.41\\ 
     \hline 
    \rowcolor{lightgray} 
    \multicolumn{7}{c}{ \textit{\textbf{Single-Image input multimodal LLMs}}} \\
    \hline 
    LLaVA-NeXT-34B~\cite{liu2024llavanext} & 22.06 & 41.03 & 38.19 & 12.00 & 38.27 & 25.00\\ 
    LLaVA-v1.5-7B-xtuner~\cite{2023xtuner} & 33.53 & 29.49 & 44.72 & 26.00 & 38.78 & 47.60\\ 
    Yi-VL-6B~\footref{Yifootnote} & 25.59 & 50.00 & 35.68 & 17.00 & 34.18 & 22.60\\ 
    LLaVA-internLM2-7B~\cite{2023internlm} & 19.12 & 39.74 & 35.43 & 12.00 & 23.98 & 28.42\\ 
    LLaVA-v1.5-13B~\cite{liu2023improvedllava} &20.00 & 25.64 & 31.66 & 20.00 & 22.96 & 20.89\\ 
    LLaVA-v1.5-7B~\cite{liu2023improvedllava} & 20.00 & 24.36 & 25.13 & 20.00 & 22.96 & 19.86\\ 
    LLaVA-v1.5-13B-xtuner~\cite{2023xtuner} & 12.94 & 30.77 & 20.10 & 11.00 & 18.37 & 21.58\\ 
    CogVLM~\cite{wang2023cogvlm} & 19.71 & 41.03 & 19.60 & 13.00 & 16.33 & 15.75\\ 
    MiniGPT-4-v2~\cite{chen2023minigptv2} & 20.00 & 21.79 & 21.61 & 13.00 & 17.35 & 14.73\\ 
    \bottomrule
    \end{tabular}}}
    \vspace{-4mm}
    \caption{\textbf{Experiment results on \BENCHMARK}. The first row shows task names and number of test data. We see that most models perform similarly to random choice, and are far from humans (\S\ref{sec:exp_analysis}).
    }
    \label{tab:main_results}
    \vspace{-5mm}
\end{table}

\subsection{Experimental Setup}
\label{sec:exp_setup}
\noindent\textbf{Multimodal LLMs: }
We evaluate \BENCHMARK on 20 recent multimodal LLMs, including models designed for considering multi-image inputs and those originally designed for single-image inputs. 
For multi-image input multimodal LLMs, we evaluate on GPT-4o, GPT-4-Turbo~\cite{gpt4}, Gemini Pro~\cite{team2023gemini}, Mantis (Idefics2, clip-llama3, and siglip-llama3 versions; 8B)~\cite{jiang2024mantis}, VILA (v1.5-13B)~\cite{lin2023vila}, Idefics (9B-Instruct and v2-8B)~\cite{laurenccon2024obelics, idefics2}, Emu2 (Chat)~\cite{sun2023generative} and OpenFlamingo (v2-9B)~\cite{awadalla2023openflamingo}.
For single-image input multimodal LLMs, we evaluate on LLaVA (v1.5, NeXT, internLM, and xtuner versions, model size 7B, 13B, and 34B)~\cite{liu2024visual, liu2023improvedllava, liu2024llavanext, 2023internlm, 2023xtuner}, Yi-VL-6B\footnote{More details are at the official website at~\url{https://www.01.ai/}\label{Yifootnote}},
MiniGPT-4-v2~\cite{chen2023minigptv2}, and CogVLM~\cite{wang2023cogvlm}.
We refer the readers to \autoref{sec:appendix:baseline} for more details.

\vspace{1ex}\noindent\textbf{Evaluation setup: }
We follow the standard setup as it is in VLMEvalKit~\cite{2023opencompass}, where the temperature is set to 0 and retry is set to 10. For the models that do not support multiple images as input, we concatenate the images to constitute one input. We extract the choice from the models' output with a set of pre-defined rules. 
We refer the readers to \Cref{sec:appendix:exp} for more details on multi-image concatenation, visual prompting, answer extraction, and the human evaluation protocol.

\subsection{Main Results}
\label{sec:exp_results}

\noindent\textbf{Overall performance: } 
As shown in \Cref{tab:main_results}, the average accuracies of the most advanced multimodal LLMs on \BENCHMARK are no better than 68\%, which are still far from enabling satisfactory utility.
The mean accuracies of open-source multimodal LLMs that have considered multi-images hover between 23.73\% and 44.50\%, which fall behind from advanced proprietary LLMs. 
Notably, there is no obvious correlation between model sizes and performances, indicating the importance of training data and training processes in developing multimodal LLMs with multi-image understanding capabilities.
For certain models and tasks, some results are only on par or even below random guessing. We provide more in-depth model analyses in the following and in \autoref{sec:appendix:errors}.

\begin{figure}[t]
    \centering
    \includegraphics[width=\textwidth]{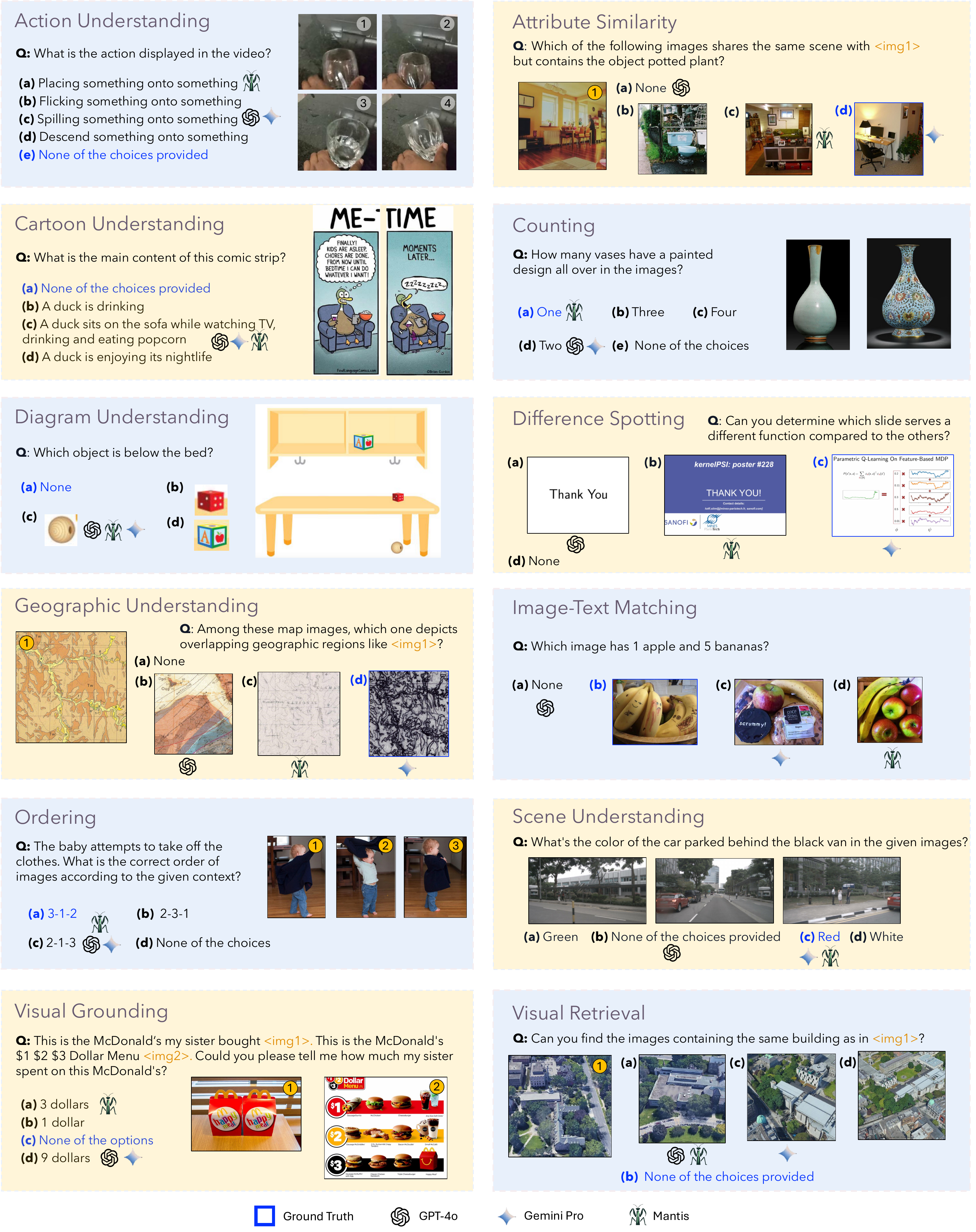}
    \caption{Qualitative results on \BENCHMARK. 
    For each task, we show the ground-truth answer in \textcolor{blue}{blue}, and choice of GPT-4o~\cite{gpt4}, Gemini Pro~\cite{team2023gemini}, and Mantis-8B-Idefics2~\cite{jiang2024mantis}. Notice that the example cases are slightly modified with change of word and image reduction for better illustration.}
    \label{fig:results-qual}
\end{figure}

\vspace{1ex}\noindent\textbf{In which multi-image tasks do multimodal LLMs show relative strengths and weaknesses?}
\Cref{fig:task} visualizes the accuracies of the best-performing models on \BENCHMARK. We observe that multimodal LLMs perform relatively better on \matching, \visualretrieval, and \diagram. In contrast, multi-image \ordering and \grounding appear to be more challenging for these models, because these tasks require understanding the whole multi-image context and conducting more complicated reasoning processes across images and modalities afterwards.

\vspace{1ex}\noindent\textbf{Can models designed for single-image inputs perform multi-image tasks?}
In general, models accepting multi-image inputs(\eg, Mantis-8B), even with fewer parameters, perform better than single-image input multimodal LLMs (\eg, LLaVA-NeXT-34B). This observation shows that generalizing from single-image training to multi-image inference is non-trivial. Reasonably, models benefit from multi-image training data and learning processes to develop multi-image understanding capabilities.

\subsection{Analysis}
\label{sec:exp_analysis}

\begin{figure}
     \centering
     \begin{minipage}[b]{0.5\textwidth}
        \centering
        \includegraphics[width=\textwidth]{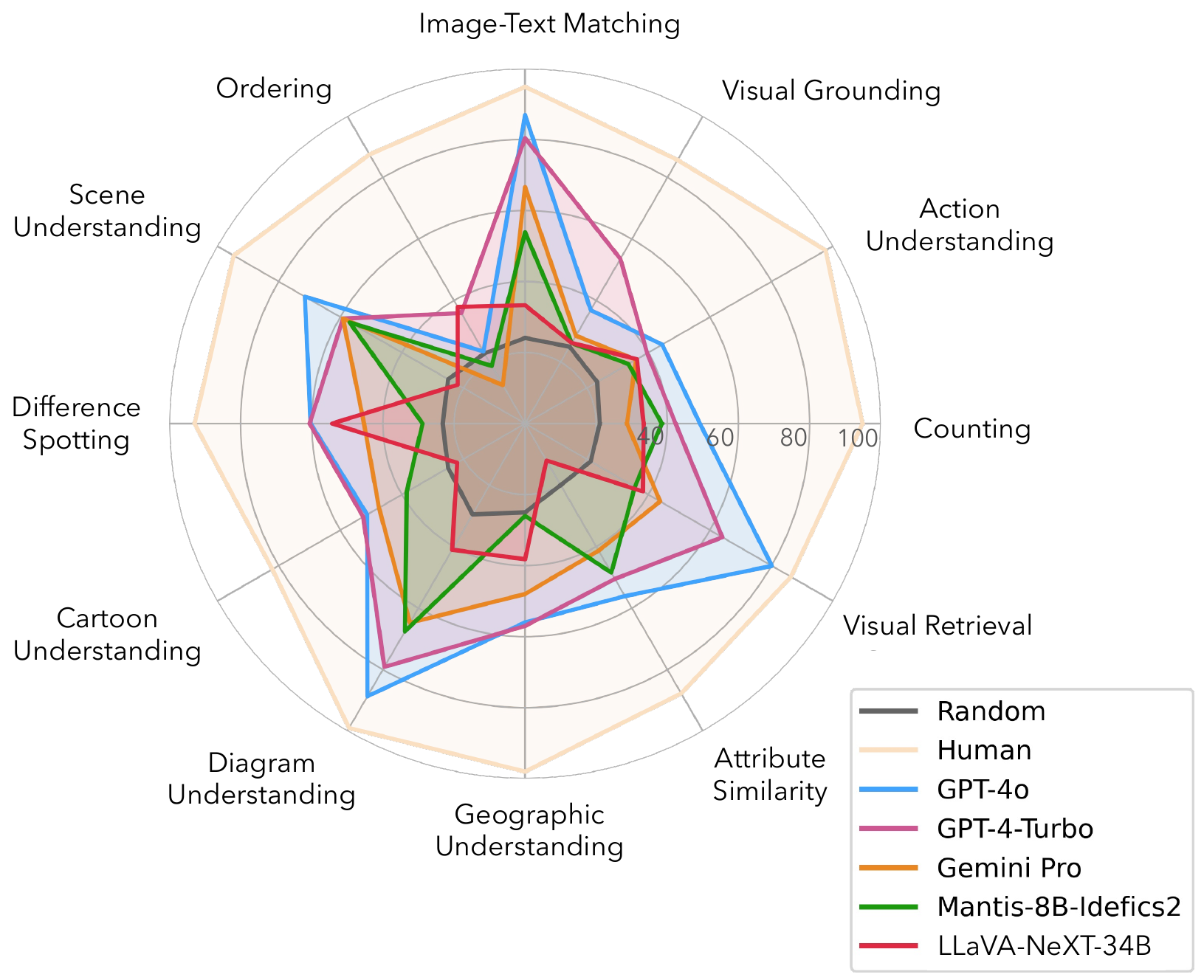}
        \vspace{-1.5em}
        \caption{Model performance by tasks.}
        \label{fig:task}
     \end{minipage}
     \hfill
     \begin{minipage}[b]{0.47\textwidth}
        \centering
        \includegraphics[width=\textwidth]{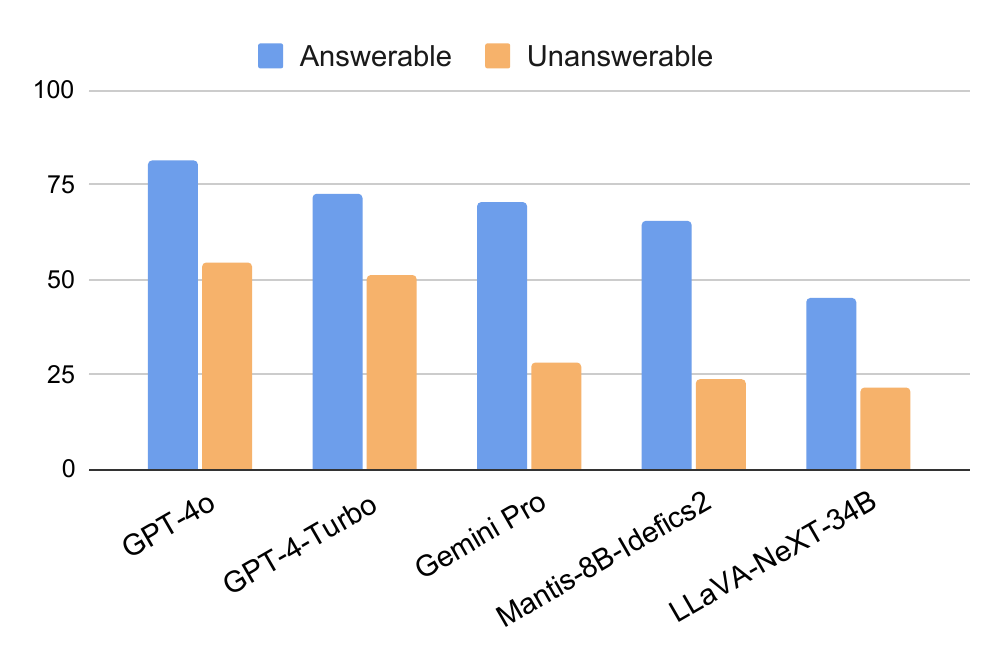}
        \vspace{-1.5em}
        \caption{Model performance on answerable and unanswerable instances. An obvious performance gap can be observed between the two sets on all best-performing models.}
        \label{fig:noanswer_result}
     \end{minipage}
     \vspace{-1em}
\end{figure}

\vspace{1ex}\noindent\textbf{Do multimodal LLMs perform worse on the unanswerable set?}
\Cref{fig:noanswer_result} compares performances on answerable and unanswerable sets for some best-performing models. All the studied models have severe performance drop when changing answerable instances to unanswerable counterparts. A closer look of the error cases reveals that models often avoid abstention when facing unanswerable questions. These observations not only highlight the importance of assessing model behavior under a more realistic setting, but also show that the pairwise design improves the reliability of \BENCHMARK.

\vspace{1ex}\noindent\textbf{Error analysis of GPT-4o:}
We randomly sampled 100 error instances made by GPT-4o on \BENCHMARK and meticulously examined them. 
The most common error category (26\% of error cases) is the failure of capturing details in images. The rest 20\% of errors are due to inaccurate object counting or reasoning, followed by errors in logical reasoning (18\%), identification of the same object in different scenes (14\%), and inferring the intents implied by image sequences (12\%).

\vspace{1ex}\noindent\textbf{Qualitative Results:}
\Cref{fig:results-qual} presents some qualitative results, one per task. A notable phenomenon is that multimodal LLMs may hallucinate by attempting to find an erroneous option that appears to be likely correct for an unanswerable question rather than abstaining (see examples for \cartoon, \diagram, \grounding, and \visualretrieval). This illustrates the obvious performance gap between answerable and unanswerable instances in \Cref{fig:noanswer_result}. We refer the readers to \Cref{sec:appendix:errors} for more in-depth discussions.

\vspace{-1mm}
\section{Conclusion}
In this work, we introduced \BENCHMARK, a comprehensive benchmark designed to provide a robust evaluation on the multi-image understanding capabilities of multimodal LLMs. Experimental results of 20 multimodal LLMs, including the prominent models like GPT-4 and Gemini Pro, revealed substantial limitations in their ability to handle multi-image scenarios. These models showed significant performance deficits compared to human accuracy and struggled more with unanswerable questions in \BENCHMARK.
Our findings underscore the need for multimodal LLMs to transcend single-image limitations and achieve more holistic visual comprehension. \BENCHMARK provides a rigorous framework for such assessments, encouraging the community to develop models that can effectively synthesize and reason across multiple visual sources.

\section*{Acknowledgments}
Special thanks to BLINK~\cite{fu2024blink} authors, especially Wei-Chiu Ma, for providing the figure templates used in this paper.
\bibliographystyle{plain}
\bibliography{ref}
\clearpage
\appendix
\addcontentsline{toc}{section}{Appendices} %
\part{Appendices} %
\parttoc %

\section{\BENCHMARK Details}
\label{sec:appendix:dataset}

\subsection{Dataset Statistics}
\begin{figure}[h]
     \centering
     \begin{minipage}[b]{0.4\textwidth}
            \centering
    \small
    \begin{tabular}{ll}
        \toprule \midrule
        Total Instances         & 2600           \\
        Total Images            & 11264          \\
        Total Tasks             & 12             \\
        Total Image Relations   & 10             \\ \midrule
        Answerable Instances    & 1300    \\
        - existing data         & 531 (40.8\%)  \\
        - derived data          & 282 (21.7\%)  \\
        - new data              & 487 (37.5\%)  \\\midrule
        Unanswerable Instances  & 1300    \\
        - change image          & 315 (24.2\%)  \\
        - change question       & 459 (35.3\%)  \\
        - change option         & 526 (40.5\%)  \\\midrule
        Average image number    & 4.3            \\
        Average question length & 21.6           \\
        Average option length   & 3.7            \\
        Average option number   & 4.4            \\ \midrule \bottomrule
    \end{tabular}
    \caption{Overall statistics of \BENCHMARK.}
    \label{table/stats}

     \end{minipage}
     \hfill
     \begin{minipage}[b]{0.58\textwidth}
        \centering
        \includegraphics[width=\textwidth]{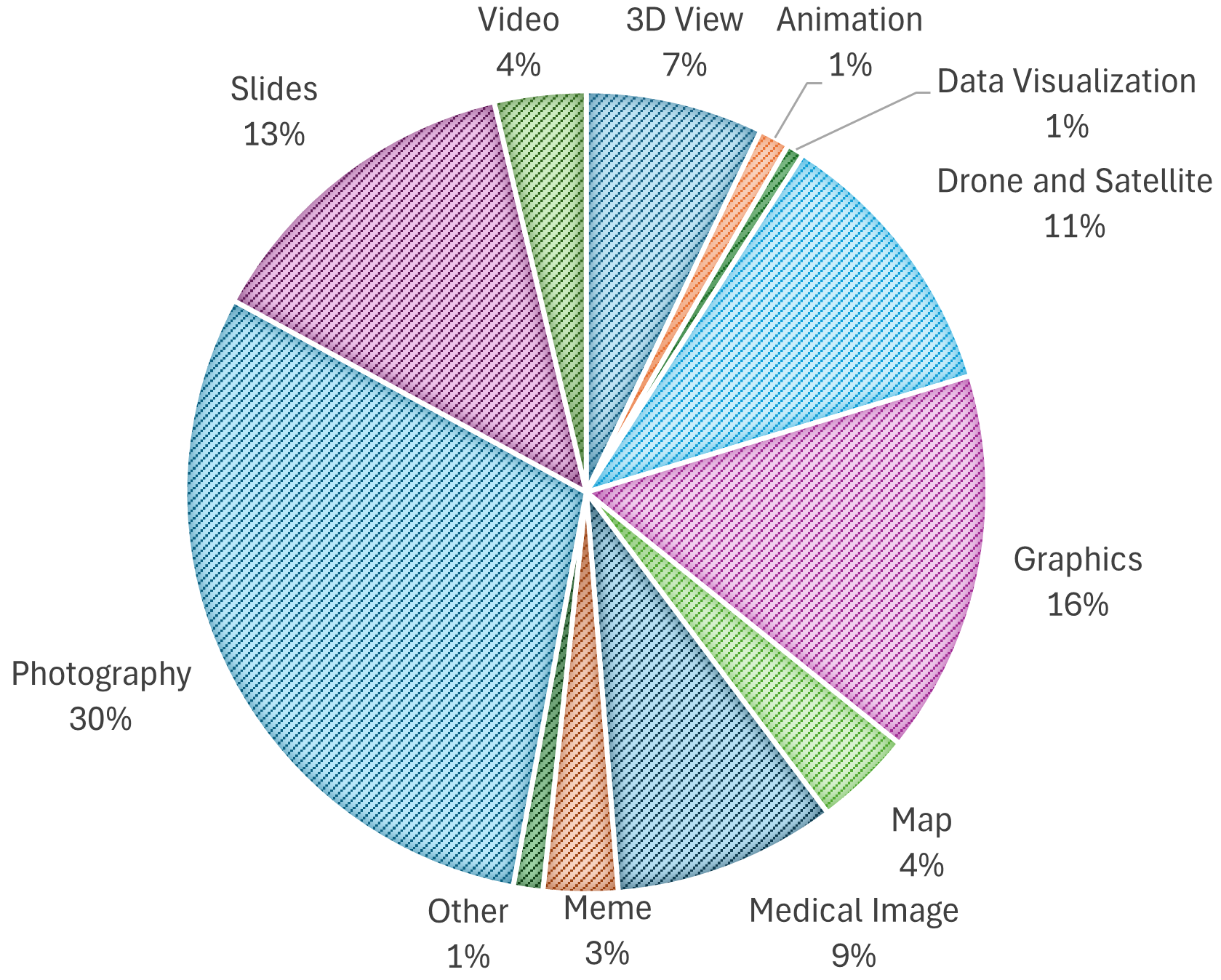}
        \caption{Data distribution by type of images.}
        \label{fig:type_dist}
     \end{minipage}
\end{figure}

\begin{figure}[h]
     \centering
     \begin{minipage}[b]{0.4\textwidth}
        \centering
        \includegraphics[width=\textwidth]{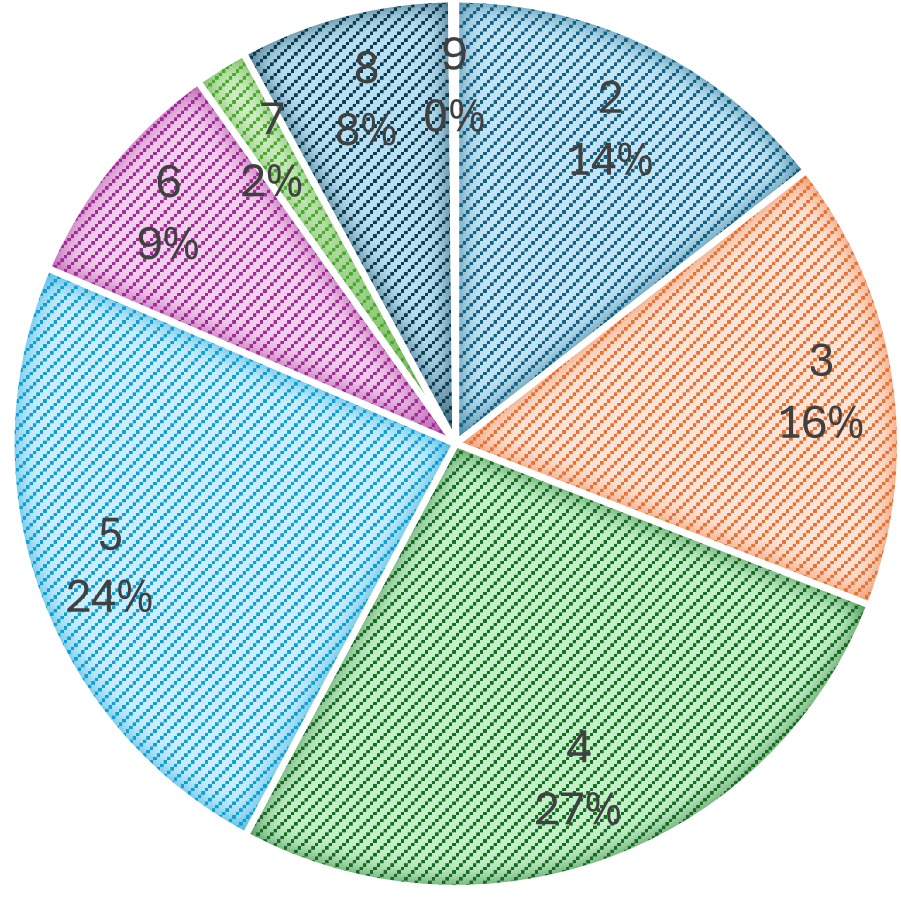}
        \caption{Data distribution by number of images.}
        \label{fig:number_dist}
     \end{minipage}
     \hfill
     \begin{minipage}[b]{0.4\textwidth}
        \centering
        \includegraphics[width=\textwidth]{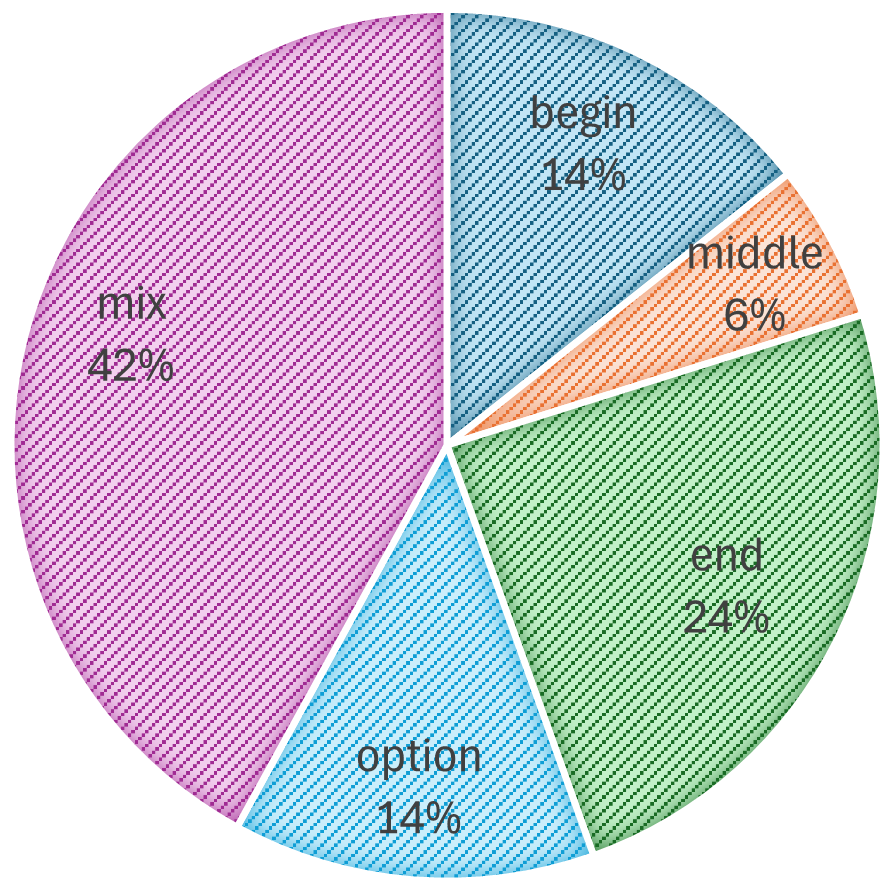}
        \caption{Data distribution by position of images.}
        \label{fig:position_dist}
     \end{minipage}
\end{figure}

\Cref{table/stats} presents the overall statistics of \BENCHMARK. \Cref{fig:type_dist} shows the data distribution by the type of images. \BENCHMARK covers a wide range of image types, ranging from common types like photography to specific areas such as medical images, slides, and drone and satellite imagery. \Cref{fig:number_dist} demonstrates the data distribution by the number of images. \BENCHMARK contains instances ranging from two images to nine images. \Cref{fig:position_dist} presents the data distribution by the position of images, including the beginning/middle/end of a question, options, and a mix of these positions.

\subsection{Dataset Curation Details}
\stitle{Answerable Data Collection}
We invest our efforts in collecting multi-image multiple-choice question answering (MCQA) data covering various tasks and multi-image relations.
Diverse data attributes enable fine-grained and diagnostic evaluation, while the multiple-choice format ensures deterministic results.
To achieve this goal, we consider three sources of data, including existing datasets, dataset derivations, as well as newly collected data. 
\textit{Existing data} come from datasets that focus on a single aspect of multi-image reasoning, such as GeneCIS \cite{vaze2023genecis}; and from datasets not specifically designed for the multi-image setting but containing a portion of multi-image data, such as SeedBench \cite{li2023seed} and IconQA \cite{lu2021iconqa}.
For a fair representation of each task, we sample up to 200 test examples from each dataset. This part contributes 40.8\% of the data in the final benchmark. 
\textit{Derived data} reformat binary QA, such as NLVR2 \cite{suhr2019corpus} and HallusionBench \cite{guan2023hallusionbench}, into MCQA by modifying questions and options; or rewriting open QA, such as ISVQA \cite{bansal2020visual}, into MCQA by adding options; and reconstructing single-image MCQA, such as MMBench \cite{liu2023mmbench}, into multi-image MCQA by replacing text options with corresponding images. Similar to those from the existing datasets, we sample up to 200 test examples from each dataset. This part contributes 21.7\% of the data in the final benchmark. 

\textit{New data} address certain tasks (\eg geographic understanding), image relations (\eg multiview), and types (\eg medical images) remaining absent or underrepresented in the aforementioned collection to fulfil a more comprehensive evaluation. We present four new datasets: HistoricalMap, UnivBuilding, PubMedMQA, and SciSlides. 
HistoricalMap requires identifying map patches covering the same regions collected from the National Geologic Map Database.\footnote{\url{https://ngmdb.usgs.gov/ngmdb/ngmdb\_home.html}} 
UnivBuilding requires identifying different views of the same building, or buildings from the same universities. The image data are from University-1652 \cite{zheng2020university, zheng2023uavm}. %
PubMedMQA contains questions regarding the subfigures from medical papers on PubMed.\footnote{\url{https://pubmed.ncbi.nlm.nih.gov/}} 
SciSlides consists of questions regarding the slides for paper presentation collected from SciDuet \cite{sun2021d2s}. 
This part contributes 37.5\% of the data in the final benchmark.

\stitle{Unanswerable Data Collection}
As shown in \Cref{fig:unanswerable}, we consider three strategies for modifying an answerable instance to its unanswerable counterpart with minimal changes. 
We first replace or reorder some images to disrupt the question-image and image-image relations. 
We also modify the question to make it incompatible with the images and options. 
In addition, we replace options to create a scenario with no correct answer. 
For each answerable instance, we apply one of these three strategies. Among all the instances, 24.2\% of the unanswerable instances are created by replacing or reordering the images in their answerable counterparts, 35.3\% by modifying the questions, and 40.5\% by changing the options.
This step doubles the size of data, leading to a balanced distribution of answerable and unanswerable instances.

\begin{figure}[t]
    \centering
    \includegraphics[width=0.9\textwidth]{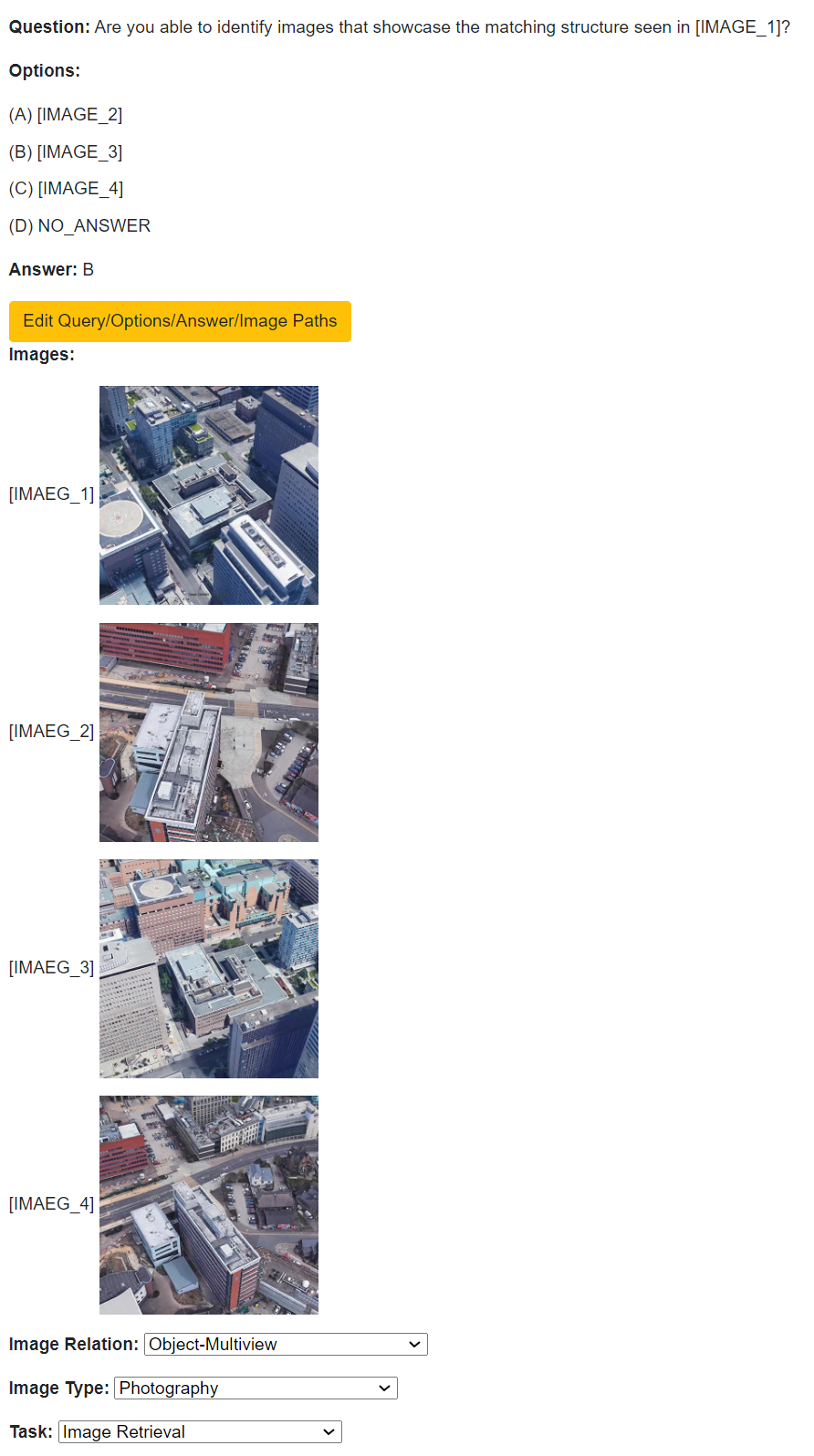}
    \caption{Annotation interface.}
    \label{fig:annotation}
\end{figure}

\stitle{Metadata Annotation}
Fine-grained metadata enable a diagnostic analysis of multimodal LLMs' weaknesses across various aspects. We annotate image relations, tasks, image types, number of images, and image positions for all instances. Among all of these attributes, image relations are a crucial factor that influences the model's capability for multi-image reasoning, yet they are rarely annotated in existing data. Therefore, we manually annotate them. Tasks and image types are partially annotated in existing data. We match the existing categories with our taxonomy and manually fill in any missing ones. Number of images and image positions are automatically detectable, so we conduct automatic annotation. The annotation interface is shown in \Cref{fig:annotation}.

\stitle{Quality Control}
We employ two types of quality control throughout the annotation process: automatic check with predefined rules, and a manual examination of each instance to filter out any low-quality data. 
The automatic check verifies valid instance format, answers, metadata values, and the coreference between image placeholders and images (ensuring no redundant image), as well as the accessibility of images.
The manual examination at last filters out ambiguous queries, unclear images, and instances with other errors, resulting in the retention of 86.3\% of instances.

\subsection{Multi-image Relations}
\BENCHMARK consists of 10 multi-image relations:
\begin{itemize}
    \item \textit{Temporal Relation}: Images are related by time, showing progression or change over a period. Examples include time-lapse photography or sequential frames from a video.
    \item \textit{Ordered Pages}: Images are part of a sequence, such as pages in a book or slides in a presentation, where the order conveys meaning.
    \item \textit{Complementary Relation}: Images that, when viewed together, provide additional information or context that enhances the understanding of the subject. They complement each other by filling in gaps or providing different perspectives.
    \item \textit{Cropped/Zoomed Images}: One image is a zoomed-in or cropped version of another, focusing on a specific part of the original image to highlight details.
    \item \textit{Narrative}: A series of images that together tell a story or convey a sequence of events, much like a comic strip or a storyboard.
    \item \textit{Scene-Multiview}: Multiple images of the same scene taken from different angles or perspectives, providing a more comprehensive view of the scene.
    \item \textit{Object-Multiview}: Images of the same object captured from various angles or perspectives, useful for understanding the object's three-dimensional shape.
    \item \textit{Overall Similarity}: Images that are generally similar in content, style, or subject matter, but not necessarily identical. They might share common themes or visual elements.
    \item \textit{Partial Similarity}: Images that share some, but not all, elements. They might have overlapping features or subjects but also contain distinct differences.
    \item \textit{Independent Images}: Images that do not have a clear relation to each other. They are not connected by time, sequence, context, or content.
\end{itemize}

\subsection{Human Evaluation Protocol}
\begin{figure}[h]
    \centering
    \includegraphics[width=0.9\textwidth]{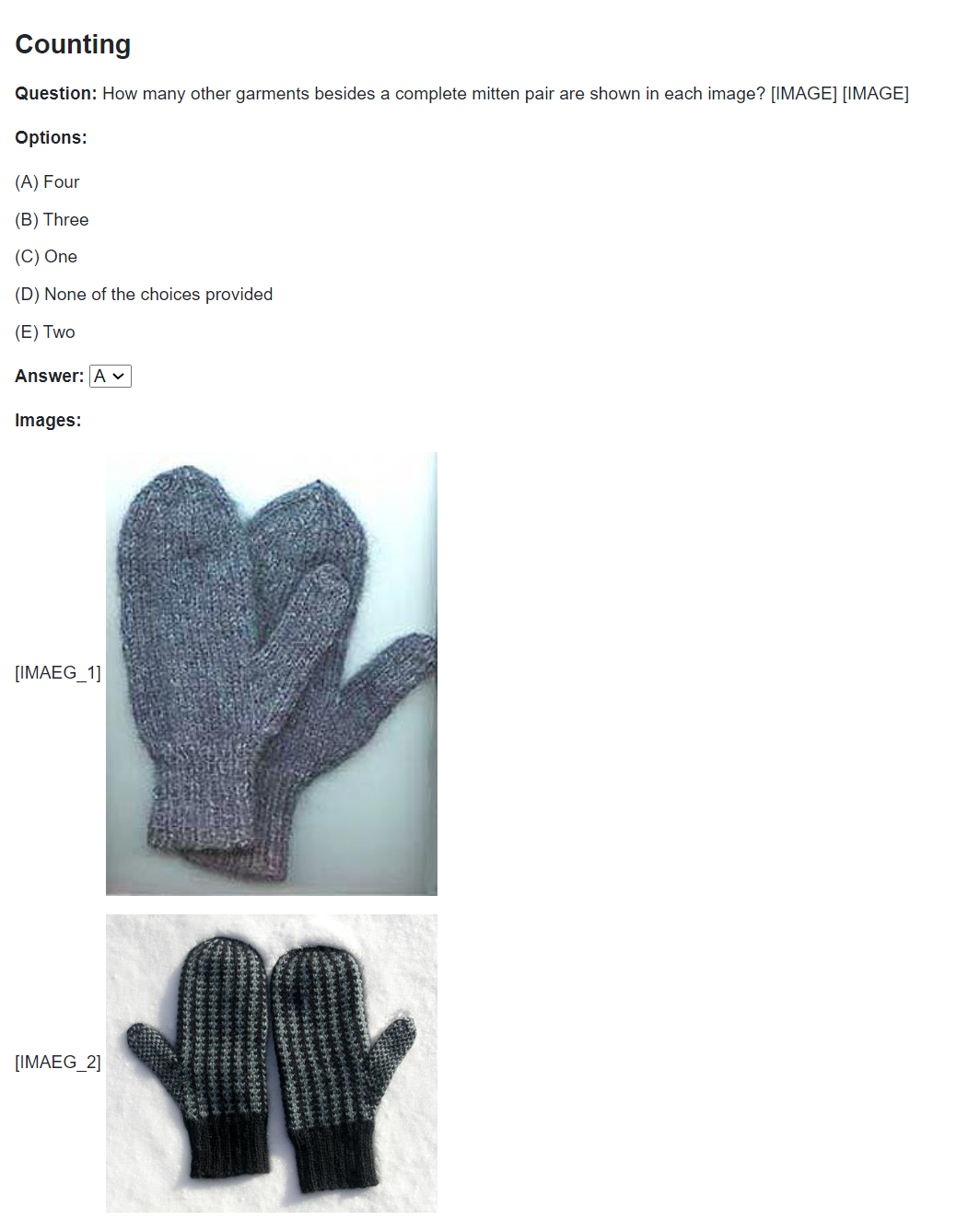}
    \caption{Human evaluation interface.}
    \label{fig:human_evaluation}
\end{figure}

Two experts in domain conduct the human evaluation. Each answerable instance and its unanswerable counterparts are randomly assigned to different experts ensuring a fair evaluation. The interface for human evaluation is shown in \Cref{fig:human_evaluation}.

\section{Baseline Models}
\label{sec:appendix:baseline}

We evaluate \BENCHMARK on 20 recent multimodal LLMs, including models designed for considering multi-image inputs and those originally designed for single-image inputs. For most model families, we use the latest and best-performing available checkpoint to date. The list of baseline models are as follows:

(i-ii) GPT-4~\cite{gpt4} is known to be one of the best multimodal models to date. We test with two most up-to-date checkpoints: gpt-4-turbo and gpt-4o. Notice that the GPT-4 performance would change if this specific checkpoint gets updated.
(iii) Gemini Pro~\cite{team2023gemini} is one of the most powerful multimodal models, and we use the Gemini 1.0 Pro Vision version of it.
(iv-vi) Mantis (Idefics2, clip-llama3, and siglip-llama3 versions; 8B)~\cite{jiang2024mantis} is a recent strong model specifically finetuned for multi-image related tasks.
(vii) VILA (v1.5-13B)~\cite{lin2023vila}, 
(viii-ix) Idefics (9B-Instruct and v2-8B)~\cite{laurenccon2024obelics, idefics2}, 
(x) Emu2 (Chat)~\cite{sun2023generative} and 
(xi) OpenFlamingo (v2-9B)~\cite{awadalla2023openflamingo} are four recent multimodal models that can take multiple images as input.
(xii-xvii) LLaVA (v1.5, NeXT, internLM, and xtuner versions, model size 7B, 13B, and 34B)~\cite{liu2024visual, liu2023improvedllava, liu2024llavanext, 2023internlm, 2023xtuner} are included as well. While they're designed for single-image input, we concatenate all the images in order.
(xviii) Yi-VL-6B\footnote{More details are at the official website at~\url{https://www.01.ai/}\label{Yifootnote}} has shown great performance recently. 
(xix) MiniGPT-4-v2~\cite{chen2023minigpt} adapts EVA~\cite{fang2023eva} as visual backbone, LLaMA2-chat (7B)~\cite{touvron2023llama} as language model backbone, and designs a linear projection layer for visual understanding abilities. 
(xx) CogVLM~\cite{wang2023cogvlm} adds a trainable visual expert module in the attention and FFN layers to bridge different modalities better. It uses EVA-CLIP~\cite{sun2023eva} as vision encoder and Vicuna~\cite{zheng2024judging} as language backbone.

\section{Experiment Setting Details}
\label{sec:appendix:exp}

\subsection{Model Prompts}
Following \cite{lu2023mathvista},\footnote{\url{https://github.com/lupantech/MathVista/blob/9ed0e8b52c0911e31faa75308082af5dcf8e63b2/evaluation/build\_query.py\#L152}} our prompt consists of four parts, the question, options, the hint indicating the answer format, and a prefix of the answer. For images, we insert them into the text to form a coherent prompt. The complete prompt is as follows:

\begin{tcolorbox}[colback=white, colframe=black!75!black, boxrule=0.5pt, sharp corners, title=Model Prompts]
\label{model_prompts}
\small
Question: \{QUESTION\}

Choices:

(A) \{OPTION\_A\}

(B) \{OPTION\_B\}

(C) \{OPTION\_C\}

(D) \{OPTION\_D\}

Hint: Please provide the correct option letter, such as A, B, C, D, directly.

Answer:
\end{tcolorbox}

\subsection{Evaluation Tool}
Following \cite{yue2023mmmu}, We use a rule-based automatic tool\footnote{\url{https://github.com/MMMU-Benchmark/MMMU/blob/f3e473e1e7af2c65a56ab66d7b3cf09c5dbaf0b9/eval/utils/eval\_utils.py\#L10}} to extract the exact answer. First, the tool detects if a valid option index appears in the model output. If no direct answer is found, the tool matches the output to the content of each option. If there is still no match, it will randomly select an option as the answer. When more than one valid answer is detected, the tool will use the first one that appears as the final answer.

\section{Error Analysis}
\label{sec:appendix:errors}

\begin{figure}[h]
     \centering
     \begin{minipage}[b]{0.49\textwidth}
        \centering
        \includegraphics[width=\textwidth]{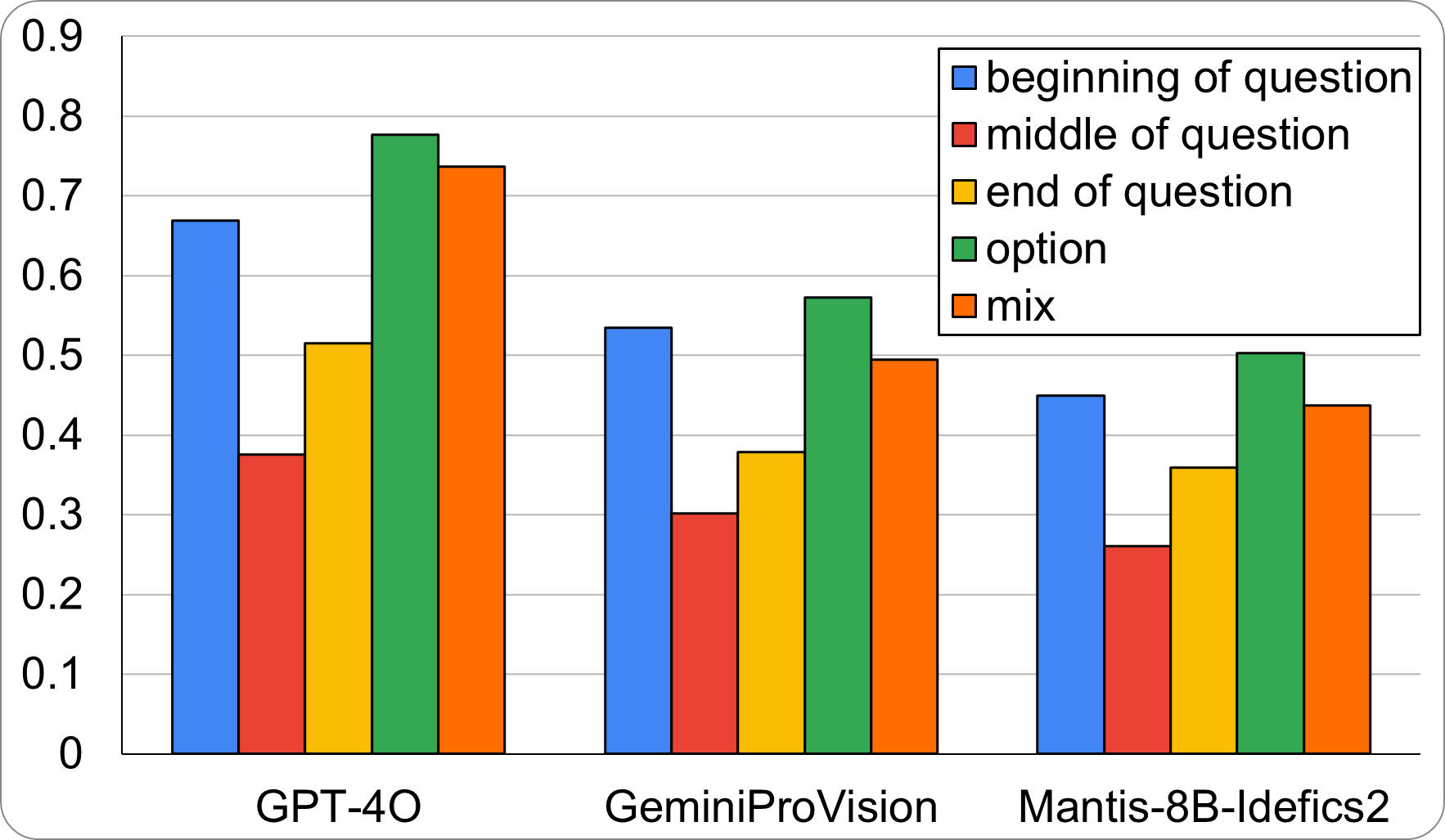}
        \caption{Model performance by image positions.}
        \label{fig:position_perf}
     \end{minipage}
     \hfill
     \begin{minipage}[b]{0.49\textwidth}
        \centering
        \includegraphics[width=\textwidth]{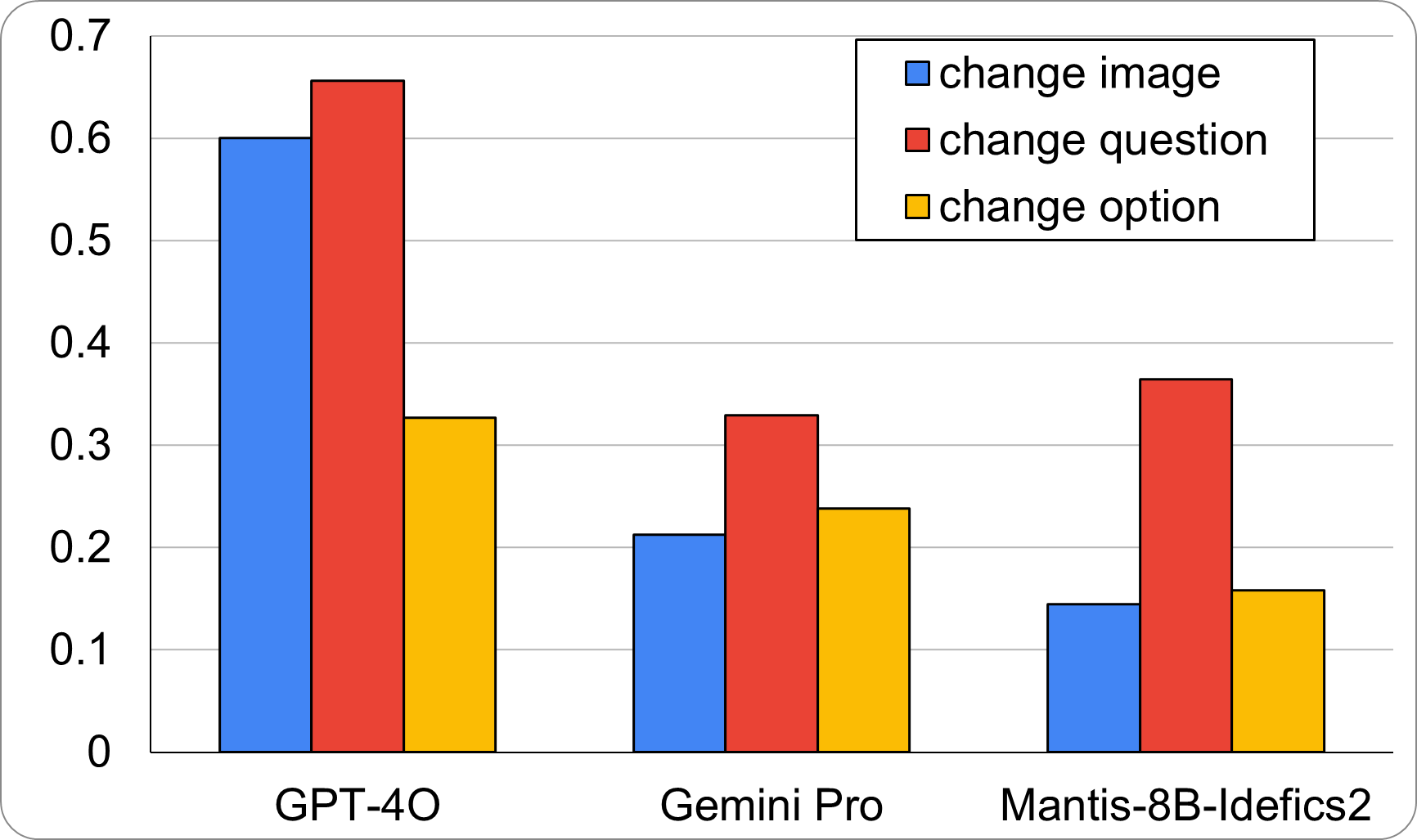}
        \caption{Model performance by unanswerable types.}
        \label{fig:unanswerable_perf}
     \end{minipage}
\end{figure}

\vspace{1ex}\noindent\textbf{Do image positions correlate with error rates?}
We analyze the error rates of varying input positions of images and report the performance of GPT-4o, GeminiProVision, and Mantis-8B-Idefics2. As shown in Figure~\ref{fig:position_perf}, the highest accuracy is achieved when images are positioned in options, while the highest error rate can be observed when images are in the middle of questions. This consistent trend across different models suggests that the position of images within a question correlates with the error rate. The cause of higher error rates might be that images in the middle or end of a question may interrupt the flow of context processing, increasing complexity and thus reducing model performance. It may also be attributed to the training process. These models may have seen less data with images in the middle during training.

\vspace{1ex}\noindent\textbf{Do unanswerable types correlate with error rates?}
We further analyze the error rates of varying unanswerable types and report the performance of the same three models in \Cref{fig:unanswerable_perf}.
Results show that the error rate also correlates with the type of unanswerable instances. All the three models perform relatively better when we only change the questions to make it incompatible with original images and options. However, all models are confused when the correct option is removed and fail to choose ``none of the other options'' in this scenario. The performance on unanswerable instances created by reordering or replacing images is divergent. GPT-4o performs much better than the other models in these cases.

\section{Limitations}
\label{sec:appendix:limitation}
\subsection{Limitation and future work}
There are several limitations to this work. First, we focus our scope on 2D images, and future research can further extend the idea of work to 3D problems, and include more multi-image tasks and relation categories. 
We hope our work can guide future efforts in providing robust and faithful evaluation in multimodal benchmarks. 
Our strategies of creating unanswerable instances, as in Figure~\ref{fig:unanswerable}, do not cover all strategies that can be used to create such instances. 
Also, we focus our evaluations on multimodal large language models. Future work could include more vision-language foundation models such as Unified-IO 2~\cite{lu2023uio2} and Chameleon~\cite{Team2024ChameleonME}.

\subsection{Societal impacts}
Our work proposes \BENCHMARK, providing a robust evaluation on multi-image tasks using multimodal LLMs. 
While it includes a comprehensive list of 12 tasks, all of them are in English and could induce bias on multilingual research settings. 
Also, if misused, the multimodal LLMs may be used to generate harmful vision and text artifacts. Nevertheless, this is not directly related to our research, and the data we curate do not contain personally identifiable information or offensive content. However, more researchers should be encouraged to get involved in research on the safety issues in a multimodal context.

\section{License}
\label{sec:appendix:license}
We release our data under CC-BY 4.0 license. For specific instances we follow their original licenses.
The datasets we used and their licenses are as follows:
\begin{itemize}
    \item \textit{GeneCIS} is released under the CC-BY-NC 4.0 license.\footnote{\url{https://github.com/facebookresearch/genecis/tree/main?tab=readme-ov-file\#license}}
    \item \textit{SEED-Bench} is released under the CC-BY-NC 4.0 license.\footnote{\url{https://huggingface.co/datasets/AILab-CVC/SEED-Bench}}
    \item \textit{IconQA} is released under the CC BY-NC-SA license.\footnote{\url{https://iconqa.github.io/}}
    \item \textit{NLVR2} is released under the CC-BY-4.0 license.\footnote{\url{https://github.com/lil-lab/nlvr/tree/master?tab=readme-ov-file\#licensing}}
    \item \textit{HallusionBench} is released under the BSD 3-Clause license.\footnote{\url{https://github.com/tianyi-lab/HallusionBench?tab=readme-ov-file\#license}}
    \item \textit{ISVQA} annotation is released under the CC BY-NC-SA 2.0 license.\footnote{\url{https://github.com/ankanbansal/ISVQA-Dataset/tree/master?tab=License-1-ov-file}} We only use the images from nuScenes, which is released under the CC BY-NC-SA 4.0 license.\footnote{\url{https://www.nuscenes.org/terms-of-use}}
    \item \textit{MMBench} is released under the Apache-2.0 license.\footnote{\url{https://github.com/open-compass/MMBench?tab=Apache-2.0-1-ov-file}}
    \item \textit{National Geologic Map Database} is free in the public domain.\footnote{\url{https://www.usgs.gov/faqs/what-are-terms-uselicensing-map-services-and-data-national-map}}
    \item \textit{University-1652} is released under the MIT license.\footnote{\url{https://github.com/layumi/University1652-Baseline?tab=MIT-1-ov-file\#readme}} %
    \item \textit{PubMed} is a free and public database, with open access articles under a Creative Commons or similar license.\footnote{\url{https://www.ncbi.nlm.nih.gov/pmc/about/copyright/}}
    \item \textit{SciDuet} is released under the Apache 2.0 license with paper slides from ACL, ICML, and NeurIPS.\footnote{\url{https://github.com/IBM/document2slides?tab=Apache-2.0-1-ov-file}}
\end{itemize}

\section{Accessibility of \BENCHMARK}

\subsection{Dataset Documentation and Format}
The full documentation of \BENCHMARK is on the project page at \url{https://huggingface.co/datasets/MUIRBENCH/MUIRBENCH}. 
For each data entry in \BENCHMARK, it includes metadata of index (idx), task, question, options, answer, image relation, image type, images, and counterpart instance idx.

\subsection{Links and Maintenance Plan}
\BENCHMARK is hosted on Huggingface/Datasets,\footnote{\url{https://huggingface.co/datasets/MUIRBENCH/MUIRBENCH}}
where license and metadata\footnote{\url{https://huggingface.co/api/datasets/MUIRBENCH/MUIRBENCH/croissant}} are also available. We maintain our benchmark on this page and will continually update it. The evaluation code and outputs will be provided to facilitate easy reproduction and analyses of the results in the paper.

\subsection{Author Statement}
We confirm that we bear all responsibility in case of violation of rights during the collection of data on \BENCHMARK, ensuring accountability and commitment to maintaining ethical standards. We will take appropriate action when needed.

\subsection{Intended Uses}
The dataset is for academic purposes only and not for commercial usage.

\end{document}